\documentclass{article}

\usepackage{microtype}
\usepackage{graphicx}
\usepackage{subcaption}
\usepackage{booktabs} 

\usepackage{hyperref}
\usepackage{url}
\usepackage{amsthm}
\usepackage{graphicx}
\usepackage{pifont}
\usepackage{bbm}
\usepackage{bm}
\usepackage{caption}
\usepackage{amsmath}
\usepackage{amssymb}
\usepackage{mathtools}
\usepackage{multirow}
\usepackage{adjustbox}
\usepackage{booktabs}
\usepackage{amssymb}
\usepackage{dirtytalk}
\usepackage{xurl}
\usepackage{thmtools}
\usepackage{thm-restate}
\usepackage{arydshln}
\usepackage{tablefootnote}
\usepackage{dirtytalk}
\usepackage{listings}
\usepackage{xcolor}

\usepackage[accepted]{icml2026}

\usepackage{amsmath}
\usepackage{amssymb}
\usepackage{mathtools}
\usepackage{amsthm}
\usepackage{colortbl}

\usepackage[capitalize,noabbrev]{cleveref}

\theoremstyle{plain}
\newtheorem{theorem}{Theorem}[section]

\newtheorem{lemma}[theorem]{Lemma}

\theoremstyle{definition}

\theoremstyle{remark}

\usepackage[textsize=tiny]{todonotes}

\icmltitlerunning{Training-Free Vector Quantization via Gaussian VAEs}

\begin{document}

\twocolumn[
  \icmltitle{Training-Free Vector Quantization via Gaussian VAEs}



  \icmlsetsymbol{equal}{*}

  \begin{icmlauthorlist}
    \icmlauthor{Tongda Xu$^*$}{air,zp,cst}
    \icmlauthor{Wendi Zheng}{zp,cst}
    \icmlauthor{Jiajun He}{ucb}
    \icmlauthor{José Miguel Hernández-Lobato}{ucb}
    \icmlauthor{Yan Wang}{air}
    \icmlauthor{Ya-Qin Zhang}{air}
    \icmlauthor{Jie Tang}{cst}
  \end{icmlauthorlist}
  \icmlaffiliation{air}{AIR, Tsinghua University}
  \icmlaffiliation{cst}{CST, Tsinghua University}
  \icmlaffiliation{zp}{Zhipu AI}
  \icmlaffiliation{ucb}{University of Cambridge}
  \icmlcorrespondingauthor{Yan Wang}{wangyan@air.tsinghua.edu.cn}
  \icmlcorrespondingauthor{Jie Tang}{jietang@tsinghua.edu.cn}
  \icmlkeywords{Vector Quantization, Variational Autoencoder}

  \vskip 0.3in
]



\printAffiliationsAndNotice{
$^*$The work was done when Tongda Xu was a full-time intern with Zhipu AI.
}  

\begin{abstract}
Vector-quantized variational autoencoders (VQ-VAEs) are discrete autoencoders that compress images into discrete tokens. However, they are difficult to train due to discretization. In this paper, we propose a simple yet effective technique dubbed \textbf{Gaussian Quant (GQ)}, which first trains a Gaussian VAE under certain constraints and then converts it into a VQ-VAE without additional training. For conversion, GQ generates random Gaussian noise as a codebook and finds the closest noise vector to the posterior mean. Theoretically, we prove that when the logarithm of the codebook size exceeds the bits-back coding rate of the Gaussian VAE, a small quantization error is guaranteed. Practically, we propose a heuristic to train Gaussian VAEs for effective conversion, named the target divergence constraint (TDC). Empirically, we show that GQ outperforms previous VQ-VAEs, such as VQGAN, FSQ, LFQ, and BSQ, on both UNet and ViT architectures. Furthermore, TDC also improves previous Gaussian VAE discretization methods, such as TokenBridge. The source code is provided in \url{https://github.com/tongdaxu/VQ-VAE-from-Gaussian-VAE}. 
\end{abstract}

\begin{figure}[thb]
\centering
    \includegraphics[width=\linewidth]{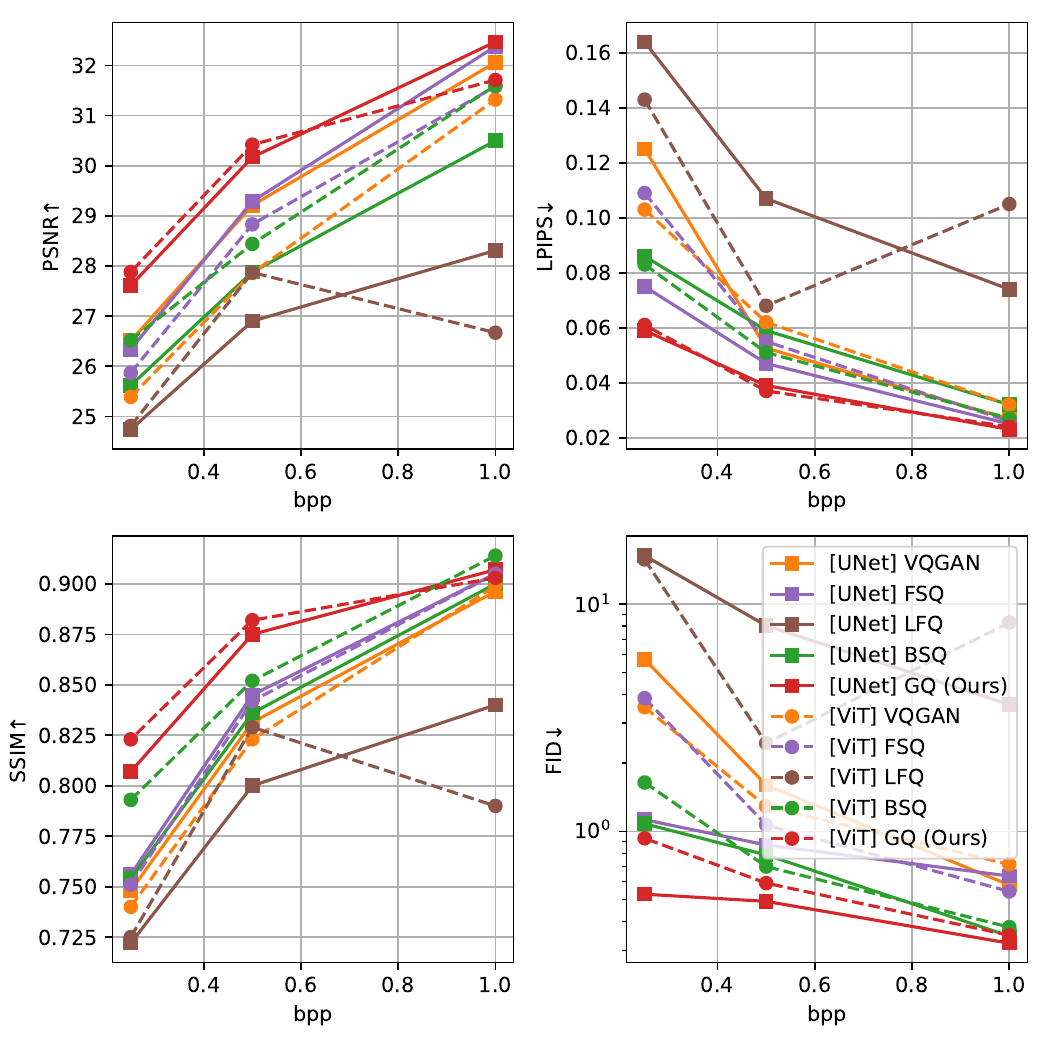}
\caption{The rate-distortion performance on the ImageNet dataset demonstrates that GQ outperforms previous VQ-VAEs on both UNet and ViT architectures. }
\label{fig:cover}
\end{figure}

\section{Introduction}

Vector-quantized variational autoencoders (VQ-VAEs) \citep{van2017neural} compress images into discrete tokens and are fundamental to autoregressive generative models \citep{esser2021taming,chang2022maskgit,yu2023language,sun2024autoregressive}. However, VQ-VAEs are difficult to train: the encoding process is non-differentiable, and issues such as codebook collapse often arise \citep{sonderby2017continuous}. As a result, special techniques are required, including commitment loss \citep{van2017neural}, expectation maximization (EM) \citep{roy2018theory}, Gumbel-Softmax \citep{jang2016categorical,maddison2016concrete,sonderby2017continuous}, and entropy loss \citep{yu2023language,zhao2024image}.

\begin{figure*}[thb]
    \centering
    \includegraphics[width=0.8\linewidth]{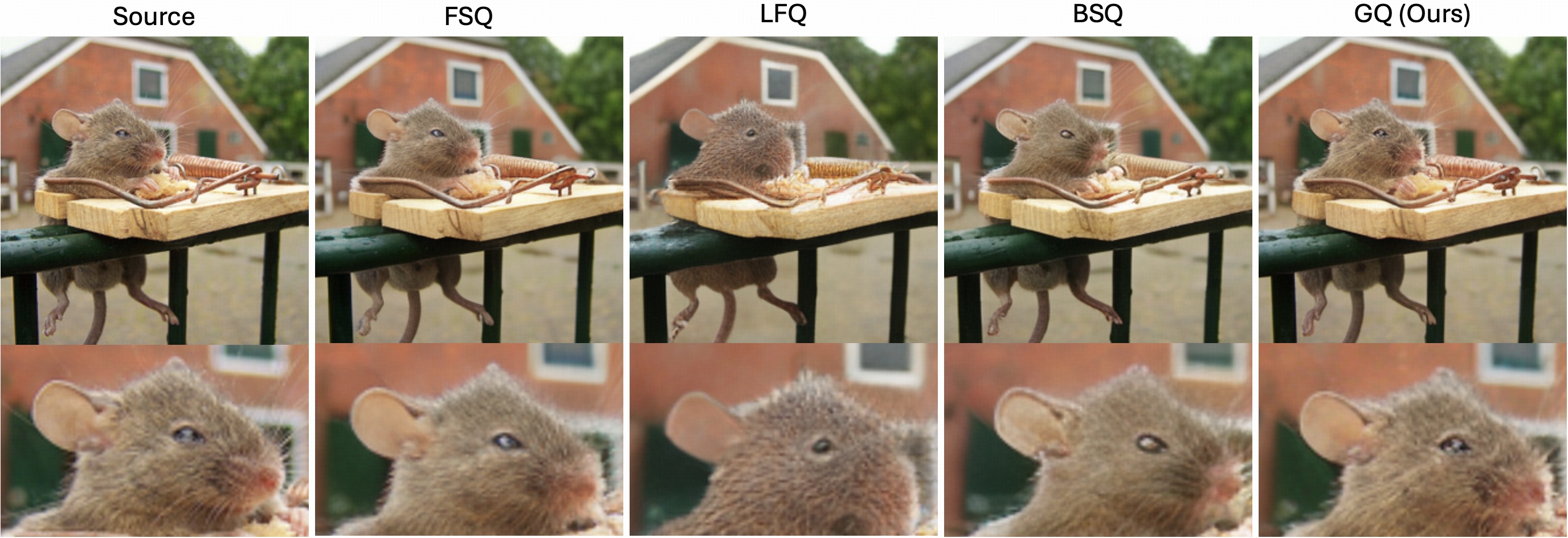}
    \includegraphics[width=0.8\linewidth]{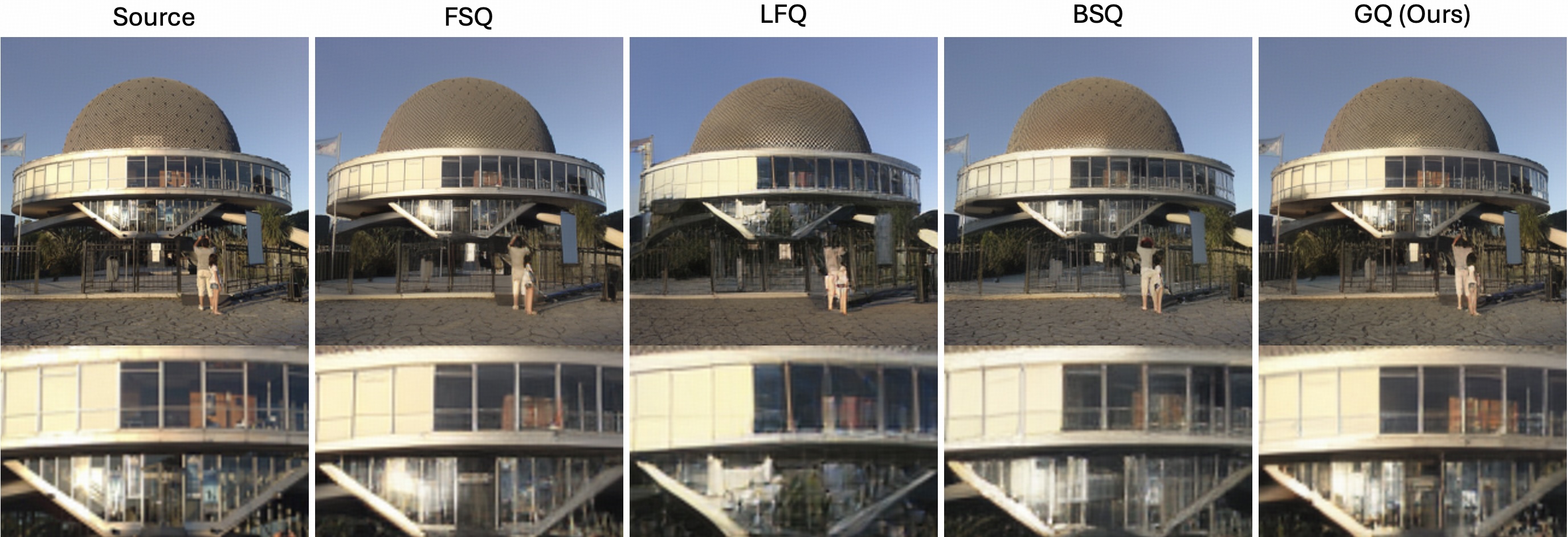}
    \caption{Qualitative results on ImageNet dataset at 0.25 bits-per-pixel (bpp). GQ has the most visually pleasing reconstruction.}
    \label{fig:qual}
\end{figure*}

In this paper, we circumvent the difficulty of training VQ-VAEs by not training them at all. Specifically, we propose \textbf{Gaussian Quant (GQ)}, which first trains a constrained Gaussian VAE and then converts it into a VQ-VAE without additional training. For conversion, GQ generates a codebook of one-dimensional Gaussian noise and, for each posterior dimension, select the entry closest to the posterior mean. We show theoretically that when the logarithm of the codebook size exceeds the bits-back coding rate \citep{hinton1993keeping,townsend2019practical} of the Gaussian VAE, the quantization error is small, providing a principled guideline for codebook size selection.

To train a Gaussian VAE for effective conversion, we introduce the \textbf{target divergence constraint (TDC)}, which encourages the Gaussian VAE to achieve the same Kullback–Leibler (KL) divergence for each dimension, corresponding to the bits-back coding bitrate. Empirically, we demonstrate that GQ with Gaussian VAE trained by TDC, outperforms previous VQ-VAEs such as VQGAN, FSQ, LFQ, and BSQ \citep{van2017neural,mentzer2023finite,yu2023language,zhao2024image} in terms of reconstruction, on UNet and ViT backbones. Additionally, we show that TDC improves previous Gaussian VAE discretization methods, such as TokenBridge \citep{wang2025bridging}.

Our contributions are summarized as follows:
\begin{itemize}
\item (Section~\ref{sec:gq}) We propose Gaussian Quant (GQ), a simple yet effective approach that constructs a VQ-VAE by training a constrained Gaussian VAE and converting it into a VQ-VAE without additional training.
\item (Section~\ref{sec:theory}) We theoretically show that when the GQ codebook size matches the bits-back coding rate of the Gaussian VAE, the conversion error remains small.
\item (Section~\ref{sec:reg}) We introduce the target divergence constraint (TDC) to train Gaussian VAEs for effective conversion, and show that GQ outperforms prior VQ-VAEs on both UNet and ViT architectures.
\item (Section~\ref{sec:ptq}) We show that TDC improves TokenBridge, an existing Gaussian VAE conversion methods.
\end{itemize}

\section{Preliminaries}
We denote the original image as $X$, the latent as $Z$, the encoder as $f(\cdot)$, and the decoder as $g(\cdot)$. We use $\log(\cdot)$ for the natural logarithm and $D_{KL}(\cdot || \cdot)$ for KL divergence. We denote the bits-back coding bitrate as $R_i = D_{KL}(q(Z_i|X)||\mathcal{N}(0,I))$. We denote the base-2 logarithm as $\log_2(\cdot)$, with the corresponding bitrate $R_i^{\mathrm{bits}} = R_i / \log 2$.

\subsection{Vector Quantized Variational Autoencoder} 
VQ-VAE \citep{van2017neural} encodes images into integer tokens. For autoregressive generation, it uses deterministic encoding and a shared codebook. Specifically, VQ-VAE maintains a codebook $c_{1:K}$ of size $K$ with bitrate $\log K$. For each dimension $i$, the encoder output $f(x)_i$ is quantized to the closest codeword $c_j$ in $c_{1:K}$. Denoting distortion as $\Delta(\cdot, \cdot)$, the VQ-VAE objective is the rate-distortion function weighted by a Lagrange multiplier $\lambda$:
\begin{gather}
    \mathcal{L}_{VQ} = \lambda \underbrace{\log K}_\textrm{bitrate} + \underbrace{\mathbb{E}[\Delta(X,g(\hat{z}))]}_{\textrm{distortion}} + \mathcal{L}_{Reg}, \notag \\
    \hat{z}_i = \arg\min_{c_j\in c_{1:K}} ||f(x)_i - c_j||,
\end{gather}
where $c_{1:K}$ is the learned codebook, and $\mathcal{L}_{\mathrm{Reg}}$ is a regularization term that ensures VQ-VAE convergence, such as commitment loss, codebook loss \citep{van2017neural}, and Gumbel-Softmax loss \citep{sonderby2017continuous}.

\subsection{Gaussian VAE and Bits-Back Coding} 
The Gaussian VAE is a special VAE \citep{kingma2013auto} with a prior $\mathcal{N}(0,I)$ and a fully factorized Gaussian posterior $q(Z | X)$. Its encoding simply involves sampling each latent dimension $z_i \sim q(Z_i|X)$. Assuming $\log p(X|Z = z) \propto (1/\lambda)\Delta(X, g(z))$, the negative evidence lower bound (ELBO) of the Gaussian VAE corresponds to a rate-distortion function, consisting of a bits-back coding bitrate term and a distortion term:
\begin{gather}
    \mathcal{L}_{VAE} = \lambda \underbrace{D_{KL}(q(Z|X)||\mathcal{N}(0,1))}_\textrm{bits-back coding bitrate} + \underbrace{\mathbb{E}[\Delta(X,g(z))]}_{\textrm{distortion}},  \notag \\
    z_i \sim q(Z_i|X=x)=\mathcal{N}(\mu_i,\sigma_i^2), i = 1 \ldots d. \label{eq:vae}
\end{gather}
The \textbf{bits-back coding bitrate} of $Z_i$ is defined as $R_i = D_{KL}(q(Z_i | X) ||\mathcal{N}(0,I))$ \citep{hinton1993keeping}, because when compressing $X$ losslessly, $z_i$ can be communicated using $R_i$ nats to arbitrary precision.

\section{Gaussian Quant: Vector Quantization using Gaussian VAE}
We propose an alternative approach to obtain a VQ-VAE: first train a constrained Gaussian VAE, then convert it into a VQ-VAE.
\subsection{Converting a Gaussian VAE into a VQ-VAE}
\label{sec:gq}
Given a Gaussian VAE, we convert it into a VQ-VAE by generating one-dimensional Gaussian noise as a codebook and quantizing the posterior mean independently for each latent dimension. The codebook is fixed once generated. Because it consists entirely of Gaussian samples, we call this approach \textbf{Gaussian Quant (GQ)}. Specifically, we randomly generate a codebook $c_{1:K} \sim \mathcal{N}(0,1)$ shared across all dimensions. Then, for each dimension $i$, we select the codeword $c_j$ closest to the posterior mean $\mu_i$ and denote the quantized value as $\hat{z}_i$:
\begin{gather}
    \hat{z}_i = \arg\min_{c_j \in c_{1:K}} ||\mu_i - c_j||,\textrm{ where } c_{1:K} \sim \mathcal{N}(0,1).\label{eq:qr}
\end{gather}
\subsection{Theoretical Relationship between the Codebook Size and Quantization Error} 
\label{sec:theory}
Why GQ works and how to choose $K$ are not obvious. Theoretically, we show that GQ preserves the rate-distortion properties of the Gaussian VAE: when the codebook bitrate $\log K$ matches the bits-back coding rate $R_i$ of the Gaussian VAE, the quantization error is small. More specifically, we denote the event $|\hat{z}_i-\mu_i|\ge\sigma_i$ as large quantization error. Then, we show that the probability of a large quantization error decays doubly exponentially as the codebook bitrate $\log K$ exceeds the bits-back coding rate $R_i$.
\begin{theorem}
\label{thm:ach}
    Denote the mean and standard deviation of $q(Z_i|X=x)$ as $\mu_i$ and $\sigma_i$, respectively. We assume $|\mu_i\sigma_i|\le c_1$ and $|\mu_i|+|\sigma_i|\le c_2$. Given a fixed $R_i = D_{KL}(q(Z_i|X)||\mathcal{N}(0,1))$, the probability of a quantization error $|\hat{z}_i - \mu_i| \geq \sigma_i$ decays doubly exponentially with the number of nats $t$ by which the codebook bitrate $\log K$ exceeds the bits-back coding rate. That is,
    \begin{gather}        
        \textup{when } \log K = R_i + t,\notag \\\Pr\{|\hat{z}_i - \mu_i| \ge \sigma_i\} \le \exp{(-e^{t}\sqrt{\frac{2}{\pi}}e^{-c_1-0.5})}.
    \end{gather}
\end{theorem}
Conversely, when the codebook bitrate $\log K$ is smaller than the bits-back coding rate $R_i$, the probability of a large quantization error increases exponentially toward $1$.
\begin{theorem}
\label{thm:con}
    Following Theorem~\ref{thm:ach}, the probability of a quantization error $|\hat{z}_i - \mu_i| \ge \sigma_i$ increases exponentially with the number of nats $t$ by which the codebook bitrate $\log K$ falls below the bits-back coding rate. That is,
    \begin{gather}        
        \textup{when } \log K = R_i - t, \notag \\ 
        \Pr\{|\hat{z}_i - \mu_i| \ge \sigma_i\} \ge 1 - e^{-t}\sqrt{\frac{2}{\pi}}e^{0.5c_2^2-0.5}.
    \end{gather}
\end{theorem}
Theorems~\ref{thm:ach} and \ref{thm:con} provide a principled guideline for choosing $K$: $\log K$ should be close to the bits-back rate $R_i$. In practice, setting $\log_2 K = \left\lceil R_i^{\mathrm{bits}} \right\rfloor$ typically yields sufficiently small reconstruction error, where $\left\lceil \cdot \right\rfloor$ denotes rounding. Using a larger $K$ offers no additional benefit, while a smaller $K$ significantly increases the error.

\subsection{Training a Conversion-ready Gaussian VAE with Target Divergence Constraint}
\label{sec:reg}
Now that we know how to convert a Gaussian VAE into a VQ-VAE, we discuss how to train a Gaussian VAE for effective conversion. To construct a VQ-VAE with a specific codebook size $K$, we need to train the Gaussian VAE so that $R_i$ is close to $\log K$ for all dimensions $i = 1, \dots, d$.

Previous works such as MIRACLE and HiFiC \citep{Flamich2020CompressingIB,Mentzer2020HighFidelityGI} have developed heuristic that limits the mean of $R_i$ across all dimensions. More specifically, they augment the original target of VAE in Eq.~\ref{eq:vae} by updating $\lambda$ according to the relationship between target bitrate $R^*$ and the mean of actual bitrate $R_i$ with update rate $\beta$:
\begin{gather}
    \mathcal{L}_{HiFiC} = \lambda \sum_{i=1}^d R_i + \mathbb{E}[\Delta(X,g(z))], \notag \\
    \lambda = \beta\lambda \textrm{ if} \textrm{ mean}_i \{R_i\} > R^* \textrm{ else } \beta^{-1}\lambda. \notag \\
\end{gather}

To train a Gaussian VAE under stricter per-dimension constraint, we propose a heuristic \textbf{Target Divergence Constraint (TDC)}, which is an extension of previous $\mathcal{L}_{HiFiC}$ heuristic with per-dimension rate control. Specifically, we set the target $R_i$ to $\log K$. For each dimension $i$, we apply a larger penalty if $R_i$ exceeds $\log K + \alpha$ and a smaller penalty if $R_i$ falls below $\log K - \alpha$, using different $\lambda$ values for each case. Here, $\alpha$ is a hyperparameter controlling the threshold:
\begin{gather}
    \mathcal{L}_{TDC} = \sum_{i=1}^d \lambda_i R_{i} + \mathbb{E}[\Delta(X,g(z))], \notag \\
    \textup{where } \lambda_i = \left\{
        \begin{array}{ll}
        \lambda_{\textrm{min}}, & R_{i}< \log K - \alpha,  \\
        \lambda_{\textrm{mean}}, & R_{i} \in [\log K\pm\alpha] , \\
        \lambda_{\textrm{max}}, & R_{i} > \log K+ \alpha .
        \end{array} \right.
        \label{eq:cvae}
\end{gather}
To determine $\lambda_{\textrm{min}}, \lambda_{\textrm{mean}}, \lambda_{\textrm{max}}$, we initialize them to 1 and update them according to the relationship between the statistics of $R_i$ and $\log K \pm \alpha$ during each gradient descent step: 
\begin{gather}
    \lambda_{\textrm{min}} = \beta\lambda_{\textrm{min}} \textrm{ if} \textrm{ min}_i \{R_i\} > \log K - \alpha \textrm{ else } \beta^{-1}\lambda_{\textrm{min}}, \notag \\
    \lambda_{\textrm{mean}} = \beta\lambda_{\textrm{mean}}  \textrm{ if} \textrm{ mean}_i 
    \{R_i\} > \log K \textrm{ else } \beta^{-1}\lambda_{\textrm{mean}}, \notag \\
    \lambda_{\textrm{max}} = \beta\lambda_{\textrm{max}}  \textrm{ if} \textrm{ max}_i\{R_{i}\} > \log K + \alpha \textrm{ else } \beta^{-1}\lambda_{\textrm{max}}, \notag 
\end{gather}
where $\beta$ is a hyperparameter controlling the update speed. To avoid numerical issues, we clip $\lambda_{\textrm{min}}, \lambda_{\textrm{mean}}, \lambda_{\textrm{max}}$ to the range $[10^{-3}, 10^3]$ after each update. 

\subsection{Supporting Multi-dimensional Codebook} 
The vanilla VQ-VAE has three key parameters: codebook size $K$, codebook dimension $m$, and number of tokens $N$. Now GQ is able to construct VQ-VAE with codebook dimension $m=1$, we can extend GQ to codebook dimension $m>1$. Specifically, given a Gaussian VAE, we can quantize $m$ latents into a single large token with codebook size
$\log_2 K = \left\lceil \sum_{k=i}^{i+m} R_k^{\mathrm{bits}} \right\rfloor$.
During quantization, we minimize the $\sigma_i$-weighted distance to find nearest neighbour, with the codebook $c_{1:K}$ sampled from $m$ dimensional factorized standard Gaussian:
\begin{gather}
    \hat{z}_{i:i+m} = \arg\min_{c_j \in c_{1:K}} ||(\mu_{i:i+m} - c_j)/\sigma_{i:i+m}||,\notag \\ \textrm{ where } c_{1:K} \sim \mathcal{N}(0,I_m).\label{eq:mqr}
\end{gather}
However, for low-bitrate cases, this vanilla $m$-dimensional conversion leads to codebook collapse. This occurs because $|\mu_i|$ is bounded by $\sqrt{2 R_i}$ (see Lemma~\ref{thm:c}), so when $R_i$ is small, codebook vectors far from $0$ are never selected. To address this, we introduce a regularization term weighted by $\omega$ that encourages the selection of $c_j$ farther from $0$:
\begin{gather}
    \hat{z}_{i:i+m} = \arg\min_{c_j \in c_{1:K}} ||(\mu_{i:i+m} - c_j)/\sigma_{i:i+m}|| - \omega||c_j||.\notag 
\end{gather}
Now we have the conversion method for multi-dimensional codebook, we consider how to train a Gaussian VAE for multi-dimensional codebook. Specifically, we modify the TDC training target to account for the relationship between the sum of $m$ bitrates $\sum_{k=i}^{i+m} R_k$ and $\log_2 K \pm \alpha$. The detailed TDC update rule is:
\begin{gather}
    \mathcal{L}_{TDC}^m = \sum_{j=1}^{d//m} \lambda_j \sum_{k=jm}^{jm+m} R_k + \mathbb{E}[\Delta(X,g(z))], \notag \\
    \textup{where } \lambda_i = \left\{
        \begin{array}{ll}
        \lambda_{\textrm{min}}, & \sum_{k=jm}^{jm+m} R_k < \log_2 K - \alpha,  \\
        \lambda_{\textrm{mean}}, & \sum_{k=jm}^{jm+m} R_k \in [\log_2 K\pm\alpha] , \\
        \lambda_{\textrm{max}}, & \sum_{k=jm}^{jm+m} R_k > \log_2 K + \alpha .
        \end{array} \right. \label{eq:tdcm} 
\end{gather}
And the update rule for $\lambda$s can be modified in a similar way:
\resizebox{0.9\linewidth}{!}{%
\begin{minipage}{\linewidth}
\begin{gather}
    \lambda_{\textrm{min}} = \beta\lambda_{\textrm{min}} \textrm{ if} \textrm{ min}_j \{\sum\nolimits_{k=jm}^{jm+m} R_k\} > \log_2K - \alpha \textrm{ else } \beta^{-1}\lambda_{\textrm{min}}, \notag \\
    \lambda_{\textrm{mean}} = \beta\lambda_{\textrm{mean}}  \textrm{ if} \textrm{ mean}_j 
    \{\sum\nolimits_{k=jm}^{jm+m} R_k\} > \log_2 K \textrm{ else } \beta^{-1}\lambda_{\textrm{mean}}, \notag \\
    \lambda_{\textrm{max}} = \beta\lambda_{\textrm{max}}  \textrm{ if} \textrm{ max}_j\{\sum\nolimits_{k=jm}^{jm+m} R_k\} > \log_2K + \alpha \textrm{ else } \beta^{-1}\lambda_{\textrm{max}}. \notag 
\end{gather}
\end{minipage}
}
\begin{table*}[thb]
\caption{Quantitative results on the ImageNet dataset, comparing different VQ methods using the same model architecture. Our GQ outperforms other VQ methods on both UNet and ViT architectures. \textbf{Bold}: best. \textcolor{gray}{Gray}: continuous Gaussian VAE.}
\label{tab:rimgnet}
\centering
\resizebox{0.9\linewidth}{!}{
\begin{tabular}{@{}lccccccccc@{}}
\toprule
\multirow{2}{*}{Method} & \multirow{2}{*}{bpp ($K\times N$)} & \multicolumn{4}{c}{UNet based} & \multicolumn{4}{c}{ViT based} \\ \cmidrule(lr){3-6}\cmidrule(lr){7-10}
 &  & PSNR$\uparrow$ & LPIPS$\downarrow$ & SSIM$\uparrow$ & rFID$\downarrow$ & PSNR$\uparrow$ &  LPIPS$\downarrow$ & SSIM$\uparrow$ & rFID$\downarrow$ \\ \midrule
  \textcolor{gray}{Gaussian VAE}  & \multirow{2}{*}{\textcolor{gray}{$\approx$ 0.25 (-)}} & \textcolor{gray}{28.60} & \textcolor{gray}{0.047} & \textcolor{gray}{0.849} & \textcolor{gray}{0.556} & \textcolor{gray}{28.95} & \textcolor{gray}{0.045} & \textcolor{gray}{0.851} & \textcolor{gray}{0.672} \\
  \textcolor{gray}{Gaussian VAE (w/ TDC)} &  & \textcolor{gray}{28.05} & \textcolor{gray}{0.053} & \textcolor{gray}{0.829} & \textcolor{gray}{0.535} & \textcolor{gray}{28.44} & \textcolor{gray}{0.054} & \textcolor{gray}{0.837} & \textcolor{gray}{0.793} \\ \arrayrulecolor{gray} \midrule \arrayrulecolor{black}
 VQGAN \citep{esser2021taming} & \multirow{5}{*}{0.25 (2$^{\textup{16}}\times$1024)} & 26.51 & 0.125 & 0.748 & 5.714 & 25.39 & 0.103 & 0.740 & 3.518 \\
 FSQ \citep{mentzer2023finite} &  & 26.34 & 0.075 & 0.756 & 1.125 & 25.87 & 0.109 & 0.751 & 3.856 \\
 LFQ \citep{yu2023language} &  & 24.74 & 0.164 & 0.722 & 16.337 & 24.81 & 0.143 & 0.725 & 15.71 \\
 BSQ \citep{zhao2024image} &  & 25.62 & 0.086 & 0.754 & 1.080 & 26.52 & 0.083 & 0.793 & 1.649 \\
 GQ (Ours)&  & \textbf{27.61} & \textbf{0.059} & \textbf{0.807} & \textbf{0.529} & \textbf{27.88} & \textbf{0.061} & \textbf{0.823}  & \textbf{0.932} \\ \midrule
 \textcolor{gray}{Gaussian VAE}  & \multirow{2}{*}{\textcolor{gray}{$\approx$ 0.50 (-)}} & \textcolor{gray}{31.18} & \textcolor{gray}{0.028} & \textcolor{gray}{0.889} & \textcolor{gray}{0.389} & \textcolor{gray}{31.64} & \textcolor{gray}{0.030} & \textcolor{gray}{0.900} & \textcolor{gray}{0.867} \\
 \textcolor{gray}{Gaussian VAE (w/ TDC)} &  & \textcolor{gray}{30.84} & \textcolor{gray}{0.035} & \textcolor{gray}{0.884} & \textcolor{gray}{0.549} & \textcolor{gray}{31.18} & \textcolor{gray}{0.033} & \textcolor{gray}{0.896} & \textcolor{gray}{0.757} \\ \arrayrulecolor{gray} \midrule \arrayrulecolor{black}
 VQGAN \citep{esser2021taming} & \multirow{5}{*}{0.50 (2$^{\textup{16}}\times$2048)} & 29.21 & 0.052 & 0.831 & 1.600 & 27.86 & 0.062 & 0.823 & 1.228 \\
 FSQ \citep{mentzer2023finite} &  & 29.29 & 0.047 & 0.845 & 0.871 & 28.83 & 0.055 &  0.842 & 1.067 \\
 LFQ \citep{yu2023language} &  & 26.90 & 0.107 & 0.800 & 8.035 & 27.87 & 0.068 & 0.829 & 2.444 \\
 BSQ \citep{zhao2024image} &  & 27.88 & 0.059 & 0.836 & 0.788 & 28.44 & 0.051 & 0.852 & 0.700 \\
 GQ (Ours)&  & \textbf{30.17} & \textbf{0.039} & \textbf{0.875} & \textbf{0.492} & \textbf{30.42} & \textbf{0.037} & \textbf{0.882} & \textbf{0.592} \\ \midrule
 \textcolor{gray}{Gaussian VAE} & \multirow{2}{*}{\textcolor{gray}{$\approx$ 1.00 (-)}} & \textcolor{gray}{32.73} & \textcolor{gray}{0.022} & \textcolor{gray}{0.910} & \textcolor{gray}{0.490} & \textcolor{gray}{32.28} & \textcolor{gray}{0.023} & \textcolor{gray}{0.910} & \textcolor{gray}{0.469} \\
 \textcolor{gray}{Gaussian VAE (w/ TDC)} &  & \textcolor{gray}{32.75} & \textcolor{gray}{0.021} & \textcolor{gray}{0.910} & \textcolor{gray}{0.294} & \textcolor{gray}{32.05} & \textcolor{gray}{0.023} & \textcolor{gray}{0.909} & \textcolor{gray}{0.397} \\ \arrayrulecolor{gray} \midrule \arrayrulecolor{black}
 VQGAN \citep{esser2021taming} & \multirow{5}{*}{1.00 (2$^{\textup{16}}\times$4096)} & 32.06 & 0.026 & 0.896 & 0.580 & 31.32 & 0.032 & 0.899 & 0.716 \\
 FSQ \citep{mentzer2023finite} &  & 32.38 & 0.025 & 0.905 & 0.636 &  31.58 & 0.026 & 0.905 & 0.544 \\
 LFQ \citep{yu2023language} &  & 28.31 & 0.074 & 0.840 & 3.617 & 26.67 & 0.105 & 0.790 & 8.288 \\
 BSQ \citep{zhao2024image} &  & 30.50 & 0.032 & 0.900 & 0.346 & 31.60 & 0.027 & \textbf{0.914} & 0.379 \\
 GQ (Ours)&  & \textbf{32.47} & \textbf{0.023} & \textbf{0.907} & \textbf{0.322} & \textbf{31.71} & \textbf{0.024} & 0.903 & \textbf{0.349} \\ \bottomrule
\end{tabular}
}
\end{table*}
\subsection{Practical Guidelines}
To summarize, GQ has the same interface as a VQ-VAE and constructs a VQ-VAE with codebook size $K$, codebook dimension $m$, and number of tokens $N$ with two steps:
\begin{itemize}
\item First, GQ trains a Gaussian VAE with latent dimension $d = mN$, using the TDC constraint with target $\log3 K$ as in Eq.~\ref{eq:tdcm}. 
\item Next, GQ converts this constrained Gaussian VAE into a VQ-VAE using Eq.~\ref{eq:mqr}. The codebook $c_{1:K}$ is fixed once generated.
\end{itemize} 

\subsection{Improving TokenBridge with Target Divergence Constraint}
\label{sec:ptq}

TokenBridge \citep{wang2025bridging} also converts a pre-trained Gaussian VAE into a VQ-VAE. It adopts the Post-Training Quantization (PTQ) technique from model compression, treating the latent as model parameters for discretization. TokenBridge uses a fixed codebook of $2^K$ centroids from a Gaussian distribution and quantizes the posterior by selecting the closest centroid. However, it directly quantizes a vanilla Gaussian VAE without constraining the KL divergence of each dimension, leading to suboptimal rate-distortion performance.

We can improve TokenBridge using TDC. Its quantization centers correspond to equal-probability partition centers of $\mathcal{N}(0,1)$, which is a special case of GQ with an evenly distributed codebook $c_{1:K}$. Therefore, the number of PTQ bits should also  match $R_i$, and TDC can enhance TokenBridge performance (see Table~\ref{tab:rptq}).

\section{Experimental Results}
\subsection{Experimental Setup}
\textbf{Models and Baselines} For image reconstruction, we use two representative autoencoder architectures: \textbf{UNet} from Stable Diffusion 3 \citep{Esser2024ScalingRF}, and \textbf{ViT} from BSQ \citep{zhao2024image}. For VQ baselines, we train  vanilla \textbf{VQGAN} \citep{esser2021taming} and several variants, including \textbf{FSQ} \citep{mentzer2023finite}, \textbf{LFQ} \citep{yu2023language}, and \textbf{BSQ} \citep{zhao2024image} on the same model architecture. Additionally, we compare to pre-trained VQ-VAEs such as \textbf{VQGAN-Taming} \citep{esser2021taming}, \textbf{VQGAN-SD} \citep{rombach2022high}, \textbf{Llama-Gen} \citep{sun2024autoregressive}, \textbf{FlowMo} \citep{Sargent2025FlowTT}, \textbf{BSQ} \citep{zhao2024image}, \textbf{OpenMagViT-V2} \citep{Luo2024OpenMAGVIT2AO}, and conversion methods like \textbf{TokenBridge} and \textbf{ReVQ}. We further show that our TDC technique improves TokenBridge \citep{wang2025bridging}, a previous training-free approach for converting Gaussian VAEs into VQ-VAEs. For image generation, we employ the Llama transformer \citep{Touvron2023Llama2O,Shi2024ScalableIT}.

\textbf{Datasets, Bitrates, and Metrics} We use the \textbf{ImageNet} \citep{Deng2009ImageNetAL} training split for training, and the \textbf{ImageNet} and \textbf{COCO} \citep{Lin2014MicrosoftCC} validation splits for testing. For reconstruction and generation experiments, images are resized to $256\times256$ and $128\times128$, respectively. For bitrates, we evaluate reconstruction performance using \textbf{codebook sizes K} from $2^{14}$ to $2^{18}$ and \textbf{number of token N} of $256$, $1024$, $2048$, and $4096$. We compute bits-per-pixel (bpp) as $\log_2 K \times N / H\times W$, where $H,W$ are image height and width. In our case, we cover bpp values of $0.07$–$1.00$. We fix channel dimension to 16 following original Stable Diffusion 3 UNet. For bpp $0.50,1.00$ we adopt codebook dimension $4,8$ respectively. For bpp $<0.50$, we adopt codebook dimension 16. For either UNet or ViT architecture, we follow ReVQ \citep{zhang2025quantizethenrectifyefficientvqvaetraining} to put codebook dimension on channel dimension. Metrics include Peak Signal-to-Noise Ratio (\textbf{PSNR}), Learned Perceptual Image Patch Similarity (\textbf{LPIPS}) \citep{Zhang2018TheUE}, Structural Similarity Index Measure (\textbf{SSIM}) \citep{Wang2004ImageQA}, and reconstruction Fréchet Inception Distance (\textbf{rFID}) \citep{Heusel2017GANsTB} for image reconstruction; and generation Fréchet Inception Distance (\textbf{gFID}) and Inception Score (\textbf{IS}) \citep{Salimans2016ImprovedTF} for image generation. Further details are provided in Appendix~\ref{app:impl}.

\begin{table*}[thb]
\caption{Quantitative results on the ImageNet dataset, compared with pre-trained VQ-VAEs. Our GQ outperforms prior pre-trained VQ-VAEs while requiring less training. \textbf{Bold}: best, $^*$: from paper, -: not available.}
\label{tab:rimgnet2}
\centering
\resizebox{\linewidth}{!}{
\begin{tabular}{@{}lccccccc@{}}
\toprule
\multirow{2}{*}{Method} & \multicolumn{1}{c}{\multirow{2}{*}{bpp ($K\times N$)}} & \multirow{2}{*}{PSNR$\uparrow$} & \multirow{2}{*}{LPIPS$\downarrow$} & \multirow{2}{*}{SSIM$\uparrow$} & \multirow{2}{*}{rFID$\downarrow$} & \multicolumn{1}{c}{\multirow{2}{*}{ImageNet Trained Epochs$\downarrow$}} & \multirow{2}{*}{Params(M)$\downarrow$} \\
& & & & & \\ \midrule
TokenBridge$^*$ \citep{wang2025bridging} & 0.375 (2$^{\textup{6}}\times$4096) & - & - & - & 1.11 & - & 83 \\
OpenMagViT-V2$^*$ \citep{Luo2024OpenMAGVIT2AO} & \multirow{3}{*}{0.07 (2$^{\textup{14}}\times$256)} & 21.63 & \textbf{0.111} & 0.640 & 1.17 & 300 & 170 \\
ReVQ-256T$^*$ \citep{zhang2025quantizethenrectifyefficientvqvaetraining} & & 21.96 & 0.121 & 0.640 & 2.05 & (Mixed dataset as DC-AE) & 83 \\ 
GQ (Ours) & & \textbf{22.30} & 0.116 & \textbf{0.642} & \textbf{1.04} & 40 & 87 \\ 
\midrule
VQGAN-Taming$^*$ \citep{esser2021taming} & \multirow{5}{*}{0.22 (2$^{\textup{14}}\times$1024)} & 23.38 & - & - & 1.190 & (OpenImages) & 67 \\
VQGAN-SD$^*$ \citep{rombach2022high} & & - & - & - & 1.140 & (OpenImages) & 83 \\
Llama-Gen-32$^*$ \citep{sun2024autoregressive} & & 24.44 & \textbf{0.064} & 0.768 & 0.590 & 40 & 70 \\ 
FlowMo-Hi$^*$ \citep{Sargent2025FlowTT} & & 24.93 & 0.073 & \textbf{0.785} & 0.560 & 300 & 945 \\ 
GQ (Ours) & & \textbf{25.31} & \textbf{0.064} & 0.762 & \textbf{0.491} & 40 & 87 \\ 
\midrule
BSQ$^*$ \citep{zhao2024image} & \multirow{2}{*}{0.28 (2$^{\textup{18}}\times$1024)} & 27.78 & 0.063 & \textbf{0.817} & 0.990 & 200 & 175 \\ 
GQ (Ours) & & \textbf{27.86} & \textbf{0.054} & 0.804 & \textbf{0.424} & 40 & 87 \\
\bottomrule
\end{tabular}
}
\end{table*}


\begin{table*}[thb]
\caption{Quantitative results of improving TokenBridge on the UNet architecture. Applying TDC significantly enhances the reconstruction quality of TokenBridge.}
\label{tab:rptq}
\centering
\resizebox{0.95\linewidth}{!}{
\begin{tabular}{@{}lccccccccc@{}}
\toprule
\multirow{2}{*}{Method} & \multicolumn{1}{c}{\multirow{2}{*}{bpp ($K\times N$)}} & \multicolumn{4}{c}{ImageNet validation} & \multicolumn{4}{c}{COCO validation} \\ \cmidrule(lr){3-6}\cmidrule(lr){7-10}
 & \multicolumn{1}{c}{} & \multicolumn{1}{c}{PSNR$\uparrow$} & \multicolumn{1}{c}{LPIPS$\downarrow$} & \multicolumn{1}{c}{SSIM$\uparrow$} & \multicolumn{1}{c}{rFID$\downarrow$} & \multicolumn{1}{c}{PSNR$\uparrow$} & \multicolumn{1}{c}{LPIPS$\downarrow$} & \multicolumn{1}{c}{SSIM$\uparrow$} & \multicolumn{1}{c}{rFID$\downarrow$} \\ \midrule
Gaussian VAE & \multirow{2}{*}{$\approx$ 1.00 (-)} & 32.73 & 0.022 & 0.910 & 0.490 & 32.64 & 0.018 & 0.917 & 2.380\\
Gaussian VAE (w/ TDC) &  & 32.61 & 0.023 & 0.906 & 0.460 & 32.69 & 0.019 & 0.919 & 2.717 \\ \midrule
TokenBridge & \multirow{3}{*}{1.00 ($2^{16}\times 4096$)} & 28.24 & 0.045 &  0.869 & 0.823 & 28.19 & 0.043 & 0.878 & 4.167 \\
TokenBridge (w/ TDC) & & 31.67 & 0.025 & 0.903& 0.385 & 31.56 & 0.022 & 0.910 & 2.171 \\
GQ (Ours) & & 32.60 & 0.022 & 0.908 & 0.280 & 32.53 & 0.020 & 0.917 & 2.153 \\ \bottomrule
\end{tabular}
}
\end{table*}
\begin{table*}[htp]
\caption{Comparison between GQ conversion and its alternatives on the ImageNet dataset. GQ achieves better reconstruction quality and can be implemented asymptotically faster when the codebook dimension $m=1$. Here, $D_{\infty}(\cdot||\cdot)$ denotes the Rényi $\infty$-divergence. }
\label{tab:rec}
\centering
\resizebox{0.9\linewidth}{!}{
\begin{tabular}{@{}lccccc@{}}
\toprule
  Methods & Encoding / Decoding Complexity &PSNR$\uparrow$ & LPIPS$\downarrow$ & SSIM$\uparrow$ & rFID$\downarrow$ \\ \midrule
Gaussian VAE (w/ TDC) & $O(1)$/$O(1)$ &32.75 & 0.023 & 0.906 & 0.460 \\ \midrule
MRC (original) & $O(2^{R_i})$/$O(1)$ & 32.09 & 0.023 & 0.906 & 0.425\\
MRC (A$^*$ coding) & $O(D_{\infty}(q(Z_i|X)||\mathcal{N}(0,1)))$/$O(D_{\infty}(q(Z_i|X)||\mathcal{N}(0,1)))$ & 32.09 & 0.023 & 0.906 & 0.425 \\
ORC  & $O(2^{R_i})$/$O(1)$ & 32.09 & 0.023 & 0.906 & 0.419 \\
GQ (Ours) & $O(R_i)$/$O(1)$ & 32.11 & 0.023 & 0.907 & 0.414 \\ \bottomrule
\end{tabular}
}
\end{table*}
\begin{table*}[htb]
\caption{Effect of TDC and alternatives. GQ is effective only when applied to a Gaussian VAE trained with the TDC constraint.}
\label{tab:con}
\centering
\resizebox{0.85\linewidth}{!}{
\begin{tabular}{@{}lcccccccc@{}}
\toprule
Methods & Constraint & $R_i^{\mathrm{bits}}$ mean, min-max & $\log_2 K$ & bpp & PSNR$\uparrow$ & LPIPS$\downarrow$ & SSIM$\uparrow$ & rFID$\downarrow$ \\ \midrule
Gaussian VAE & None & 3.99, 0.26-27.29 & - & 1.00 & 32.73 & 0.022 & 0.910 & 0.490 \\
GQ & - & - & 4 & 1.00 & 26.43 & 0.054 & 0.834 & 0.978 \\ \midrule
Gaussian VAE & MIRACLE / HiFiC & 4.34,0.91-26.98 & - & 1.00 & 32.82 & 0.023 & 0.910 & 0.436 \\
GQ & - & - & 4 & 1.00 & 29.48 & 0.039 & 0.887 & 0.439 \\ \midrule
Gaussian VAE & IsoKL &  4.34, 4.24-4.38 & - & 1.00 & 30.54 & 0.878 & 0.027 & 0.400 \\
GQ & - & - & 4 & 1.00 & 30.45 & 0.878 & 0.030 & 0.468 \\ \midrule
Gaussian VAE & TDC (Ours, m=1) &  4.26, 2.93-5.63 & - & 1.06 & 32.61 & 0.023 & 0.906 & 0.460 \\
GQ & - &  -  & 4 & 1.00 & 32.11 & 0.023 & 0.906 & 0.414 \\ 
\bottomrule
\end{tabular}
}
\end{table*}

\subsection{Main Results}
\textbf{Image Reconstruction} Tables~\ref{tab:rimgnet} and~\ref{tab:rococo} compare GQ with other quantization methods with same model architecture. Across reconstruction metrics such as PSNR, LPIPS, SSIM, and rFID, GQ achieves state-of-the-art performance in most cases. Its advantage is consistent across UNet and ViT architectures, as well as on the ImageNet and COCO datasets. Moreover, the quantization brought by GQ is marginal compared to Gaussian VAE. Figure~\ref{fig:qual} shows that GQ produces visually pleasing reconstructions, preserving substantially more details from the source image. Moreover, Table~\ref{tab:rimgnet2} shows that GQ achieves competitive performance relative to several pre-trained models, such as FlowMo, while requiring fewer training epochs and smaller or comparable model size. Finally, GQ outperforms prior Gaussian VAE discretization methods, including TokenBridge and ReVQ.

\textbf{Improving TokenBridge} Table~\ref{tab:rptq} compares TokenBridge \citep{wang2025bridging} applied to a vanilla Gaussian VAE versus a TDC-constrained Gaussian VAE. The results show that TokenBridge incurs substantial quantization error on a vanilla Gaussian VAE, whereas applying TDC significantly reduces this error.

\begin{table}[thb]
\caption{Quantitative results for class-conditional image generation on the ImageNet dataset. GQ achieves the highest codebook utilization while maintaining competitive generation performance. \textbf{Bold}: best. \underline{Underline}: second best.}
\label{tab:rgen}
\centering
\resizebox{\linewidth}{!}{
\begin{tabular}{@{}lcccc@{}}
\toprule
Method & Codebook usage$\uparrow$ & gFID $\downarrow$ & IS $\uparrow$ \\ \midrule
\textit{Diffusion} &  &  &  \\
Gaussian VAE w/o TDC & - & 8.35 & 202.19  \\
Gaussian VAE w/ TDC & - & 8.47 & 205.94 \\ \midrule
\textit{Auto-regressive} &  &  &  \\
VQGAN & 16.4\%  & 8.01 & 151.40 \\
FSQ & 94.3\%  & \textbf{7.33} & \underline{224.88} \\
LFQ & 24.9\% & 7.73 & 142.09 \\
BSQ & \underline{99.8\%}  & 7.82 & 221.64 \\
TokenBridge & 94.6\% & 7.82 & 198.24 \\
GQ (Ours) & \textbf{100.0\%}& \underline{7.67} & \textbf{230.79} \\ \bottomrule
\end{tabular}
}
\end{table}

\begin{table}[thb]
\caption{The reconstruction result when $\log K\neq R_i$. GQ works best when $\log K$ matches $R_i$.}
\label{tab:sics}
\centering
\resizebox{\linewidth}{!}{
\begin{tabular}{@{}ccccccc@{}}
\toprule
$R_i^{\mathrm{bits}}$ & $\log_2 K$ & $K\times N$ & PSNR$\uparrow$ & LPIPS$\downarrow$ & SSIM$\uparrow$ & rFID$\downarrow$ \\ \midrule
14 & 14 & $2^{14}\times 1024$ & 25.31 & 0.064 & 0.762 & 0.491 \\
18 & 14 & $2^{14}\times 1024$ & 25.24 & 0.068 & 0.774 & 0.527 \\
14 & 18 & $2^{18}\times 1024$ & 27.79 & 0.059 & 0.808 & 0.513 \\
18 & 18 & $2^{18}\times 1024$ & 27.86 & 0.054 & 0.804 & 0.424 \\ \bottomrule
\end{tabular}
}
\end{table}

\begin{table}[thb]
\caption{GQ is robust to random codebook.}
\label{tab:seed}
\centering
\resizebox{0.78\linewidth}{!}{
\begin{tabular}{@{}ccccc@{}}
\toprule
Random Seed & PSNR$\uparrow$ & LPIPS$\downarrow$ & SSIM$\uparrow$ & rFID$\downarrow$ \\ \midrule
42 & 27.61 & 0.059 & 0.807 & 0.529 \\
43 & 27.61 & 0.059 & 0.807 & 0.523 \\
44 & 27.62 & 0.059 & 0.807 & 0.526 \\ \bottomrule
\end{tabular}
}
\end{table}

\begin{table}[thb]
\caption{The necessity of GQ’s two-stage pipeline is demonstrated: converting a pre-trained Gaussian VAE yields better performance than training GQ from scratch, and further fine-tuning provides only marginal improvements.}
\label{tab:ablgvae}
\centering
\resizebox{\linewidth}{!}{
\begin{tabular}{@{}lccccc@{}}
\toprule
Method & bpp & PSNR$\uparrow$ & LPIPS$\downarrow$ & SSIM$\uparrow$ & rFID$\downarrow$ \\ \midrule
 VQ-VAE from scratch & \multirow{4}{*}{1.00} & 8.50 & 0.763 & 0.156 & 360 \\
Eq.~\ref{eq:mqr} from scratch & & 29.65 & 0.044 & 0.866 & 0.928 \\
Convert from Eq.~\ref{eq:mqr} &  &  32.47 & 0.023 & 0.907 & 0.327 \\
Further finetune Eq.~\ref{eq:mqr} &  & 32.45  & 0.022 & 0.905 & 0.264 \\
 \bottomrule
\end{tabular}
}
\end{table}

\begin{table}[thb]
\caption{Ablation study on hyperparameters $\alpha$, $\beta$, and codebook dimension $m$. Here, $\alpha^{\mathrm{bits}} = \alpha / \log 2$.}
\label{tab:tdcp}
\centering
\resizebox{0.9\linewidth}{!}{
\begin{tabular}{@{}ccccccc@{}}
\toprule
$\alpha^{bits}$ & $\beta$ & $m$ & PSNR$\uparrow$ & LPIPS$\downarrow$ & SSIM$\uparrow$ & rFID$\downarrow$ \\ \midrule
0.5 & 1.01 & 1 & 25.60 & 0.088 & 0.702 & 1.26 \\ 
0.5 & 1.01 & 4 & 26.98 & 0.068 & 0.793 & 0.927 \\ 
0.5 & 1.01 & 16 & 27.61 & 0.059 & 0.807 & 0.529 \\
0.1 & 1.01 & 16 & 27.56 & 0.058 & 0.812 & 0.551 \\
1.0 & 1.01 & 16 & 27.61 & 0.063 & 0.811 & 0.701 \\
0.5 & 1.1 & 16 & 27.63 & 0.060 & 0.809 & 0.534 \\
0.5 & 1.001 & 16 & 27.48 & 0.058 & 0.804 & 0.510 \\
\bottomrule
\end{tabular}
}
\end{table}

\textbf{Image Generation} Table~\ref{tab:rgen} evaluates GQ for image generation. Compared with VQGAN, FSQ, LFQ, and BSQ, GQ achieves higher codebook usage and codebook entropy. In terms of generation FID and IS, GQ is comparable to FSQ and outperforms the other methods. Additionally, we train a DiT \citep{Peebles2022ScalableDM} with the same architecture and training setup using Gaussian VAEs with and without TDC. The results show that, under limited computational resources, autoregressive generation is more efficient than diffusion in both FID and IS. This demonstrates that converting a Gaussian VAE to a VQ-VAE facilitates autoregressive generation, improving image generation efficiency.

\textbf{Complexity} Compared with a Gaussian VAE, the computational overhead of GQ is negligible (see Appendix~\ref{app:complex}).

\subsection{Ablation Study}
\label{sec:abl}

\textbf{GQ Conversion: Alternatives} Stochastic alternatives to GQ conversion exist, including MRC, ORC, and A$^*$ coding \citep{havasi2018minimal,Theis2021AlgorithmsFT,Flamich2022FastRE,he2024accelerating}. Table~\ref{tab:rec} compares these methods in terms of reconstruction quality. When applied to a TDC-constrained Gaussian VAE, GQ achieves the best PSNR, SSIM, and rFID. Moreover, for codebook dimension $m=1$, GQ can be implemented via bisection search, making it asymptotically faster (see Appendix~\ref{app:asycomplex}).

\textbf{GQ Conversion: Selection of Codebook Size} Theorem~\ref{thm:ach}-\ref{thm:con} show that setting codebook size $\log K = R_i$ is theoretically optimal. In Table~\ref{tab:sics}, we show that empirically GQ is most effective when $\log K$ matches $R_i$.

\textbf{GQ Conversion: Robustness to Random Codebook} The codebook $c_{1:K}$ is sampled from a Gaussian distribution and fixed once generated. Table~\ref{tab:seed} shows that GQ’s performance is nearly invariant to different random seeds.

\textbf{GQ Training: TDC} To evaluate the necessity of TDC in Eq.~\ref{eq:cvae}, we train a vanilla Gaussian VAE without TDC. As shown in Table~\ref{tab:con}, the mean $R_i^{\mathrm{bits}}$ of the vanilla VAE is close to that of the TDC-constrained VAE (3.99 vs. 4.26 bits), but the range of $R_i^{\mathrm{bits}}$ is much wider for the vanilla model (0.26–27.29 vs. 2.93–5.63 bits). While the reconstruction performance of the two Gaussian VAEs is similar (PSNR: 32.73 vs. 32.61 dB, rFID: 0.490 vs. 0.460), GQ applied to the TDC-constrained VAE significantly outperforms GQ applied to the vanilla VAE (PSNR: 31.25 vs. 26.43 dB, rFID: 0.372 vs. 0.978). This demonstrates that TDC is essential for effective GQ conversion.

Alternatives to TDC include the MIRACLE / HiFiC heuristic \citep{Havasi2018MinimalRC, Mentzer2020HighFidelityGI} and IsoKL \citep{Flamich2022FastRE}. MIRACLE / HiFiC is less effective at controlling the range and is outperformed by TDC (PSNR 29.48 vs. 32.11 dB). IsoKL imposes a stricter constraint by requiring that $R_i$ is identical across all dimensions. While IsoKL enforces this constraint effectively, its overall performance is worse (PSNR 30.45 vs. 32.11 dB) due to numerical instability and the exclusion of solutions with $\sigma_i^2 > 1$. In Appendix~\ref{app:smkl}, we propose a numerically stable version of Mean-KL \citep{Lin2023MinimalRC}, which extends IsoKL to support multi-dimensional codebooks ($m > 1$). However, it performs poorly for ViT-based models.

\textbf{Necessity of GQ's Two-Stage Pipeline} One could train a vanilla VQ-VAE \citep{van2017neural} using the same codebook as GQ, effectively creating a VQ-VAE with a fixed Gaussian noise codebook. Alternatively, the Gaussian VAE network could be trained directly with the GQ target in Eq.~\ref{eq:mqr} using Gumbel-Softmax \citep{jang2016categorical,maddison2016concrete}. However, as shown in Table~\ref{tab:ablgvae}, both approaches fail to converge reliably. Moreover, fine-tuning GQ after initializing with a pre-trained Gaussian VAE yields only marginal performance improvements.

\textbf{Hyperparameters} To justify our choice of TDC parameters $\alpha = 0.5$ and $\beta = 1.01$, we perform a grid search with results shown in Table~\ref{tab:tdcp}. The results indicate that the effect of $\alpha$ and $\beta$ on TDC is minimal as long as $\alpha \le 0.5$. Table~\ref{tab:tdcp} also shows that a larger codebook dimension $m$ improves reconstruction. Accordingly, we set $m$ to the quotient of the total latent dimension and the total number of tokens, which is the maximum feasible value. To achieve optimal reconstruction, we use $\omega = 2.0$ for bitrates $\le 0.50$ bpp and $\omega = 0.0$ for 1.00 bpp (see Table~\ref{tab:ablbeta} for details).

\section{Related works}

\textbf{Vector-Quantized Variational Autoencoder} VQ-VAE \citep{van2017neural} is an autoencoder that compresses images into discrete tokens. Due to discretization, it cannot be trained directly via gradient descent. Various techniques address this, including commitment loss \citep{van2017neural}, expectation maximization (EM) \citep{roy2018theory}, the straight-through estimator (STE) \citep{bengio2013estimating}, and Gumbel-Softmax \citep{jang2016categorical,maddison2016concrete,sonderby2017continuous,Shi2024TamingSV}. VQ-VAE is also prone to codebook collapse. To mitigate this, methods such as reducing code dimension \citep{Yu2021VectorquantizedIM,Sun2024AutoregressiveMB}, product quantization \citep{Zheng2022MoVQMQ}, residual quantization \citep{Lee2022AutoregressiveIG}, dynamic quantization \citep{Huang2023TowardsAI}, multi-level quantization \citep{Razavi2019GeneratingDH}, feature projection \citep{Zhu2024ScalingTC}, and rotation codebooks \citep{Fifty2024RestructuringVQ} have been proposed \citep{yu2021vector,chiu2022self,takida2022sq,zhang2023regularized,huh2023straightening,gautam2023soft,goswami2024hypervq}.

More closely related to our work, some VQ-VAE variants have fixed codebooks, including FSQ \citep{mentzer2023finite}, LFQ \citep{yu2023language}, BSQ \citep{zhao2024image} and their extensions \citep{Zhuang2025WeTokPD,Lin2026iFSQIF}. However, training still relies on tricks such as the straight-through estimator (STE). On the other hand, TokenBridge \citep{wang2025bridging} and ReVQ \citep{zhang2025quantizethenrectifyefficientvqvaetraining} convert a pre-trained Gaussian VAE into a VQ-VAE, but they do not constrain the divergence of the Gaussian VAE.

\textbf{Reverse Channel Coding} \label{sec:rec}
GQ is closely related to reverse channel coding, which aims to simulate a distribution $q$ using samples from another distribution $p$ \citep{harsha2007communication,li2018strong,havasi2018minimal,Flamich2020CompressingIB,Theis2021AlgorithmsFT,Flamich2022FastRE,he2024accelerating}. The key difference between MRC and GQ is that MRC and its variants \citep{havasi2018minimal,Theis2021AlgorithmsFT,Flamich2022FastRE,he2024accelerating} simulate a distribution via stochastic sampling, whereas a VQ-VAE requires deterministic quantization. Regarding quantization error, GQ outperforms MRC by construction (Eq.~\ref{eq:qr}). Moreover, GQ with a one-dimensional codebook ($m=1$) can be implemented via bisection search, achieving superior asymptotic complexity (see Appendix~\ref{app:asycomplex}).

Additionally, TDC is closely related to the MIRACLE / HiFiC heuristic and IsoKL parameterization of Gaussian VAEs \citep{Havasi2018MinimalRC,Mentzer2020HighFidelityGI,Flamich2022FastRE,Lin2023MinimalRC}. Specifically, MIRACLE / HiFiC also adjusts $\lambda$ during VAE training, but it maintains a single $\lambda$, making it less effective at controlling the minimum and maximum values of $R_i$. IsoKL, in contrast, enforces strict control over $R_i$ by directly solving for $\sigma$ given $\mu$ using the Lambert $\mathcal{W}$ function \citep{corless1996lambert,brezinski1996extrapolation}, but it suffers from numerical instability and yields suboptimal performance.

\section{Conclusion \& Discussion}

In this paper, we propose \textbf{Gaussian Quant (GQ)}, a simple yet effective method that constructs a VQ-VAE by first training a constrained Gaussian VAE then converting it into a VQ-VAE. The conversion process involves generating codebook from pure Gaussian noise  and matching the posterior mean. Theoretically, we show that when the logarithm of the GQ codebook size exceeds the bits-back coding bitrate of the Gaussian VAE, the resulting quantization error is small. To train a constrained Gaussian VAE for effective conversion, we introduce the target divergence constraint (TDC). Empirically, GQ outperforms prior discrete VAEs, including VQGAN, FSQ, LFQ, and BSQ \citep{van2017neural,mentzer2023finite,yu2023language,zhao2024image}, and TDC further improves the performance of TokenBridge \citep{wang2025bridging}.

One disadvantage of GQ is that it is more complex to implement and contains additional hyper-parameter on top of VQ. However, it is noteworthy that GQ also eliminates the need of STE, codebook loss, commitment loss and entropy loss. We acknowledge that several highly competitive VQ-VAEs employ multi-scale or residual architectures \citep{Razavi2019GeneratingDH,Lee2022AutoregressiveIG,Tian2024VisualAM,Han2024InfinitySB}. In this work, however, we use a standard single-scale architecture to focus on the core aspects of the quantization method. Furthermore, recent studies show that strong reconstruction does not necessarily imply strong generation, and achieving good generation often requires feature alignment \citep{wang2024image,xiong2025gigatok,hansen2025learnings}. Here, we focus on reconstruction and leave the exploration of the complex relationship between reconstruction and generation performance to future work.

\section*{Impact Statement}
The approach proposed in this paper focus on reconstruction of existing images with limited bitrate. As the model is essentially not generative, the ethic concerns is not obvious. Nevertheless, the GAN module in decoder might has negative effects, including issues related to mis-representation and trustworthiness.

\bibliography{example_paper}

@misc{kingma2013auto,
  title={Auto-encoding variational bayes},
  author={Kingma, Diederik P and Welling, Max and others},
  year={2013},
  publisher={Banff, Canada}
}

@article{jang2016categorical,
  title={Categorical reparameterization with gumbel-softmax},
  author={Jang, Eric and Gu, Shixiang and Poole, Ben},
  journal={arXiv preprint arXiv:1611.01144},
  year={2016}
}

@article{Zheng2022MoVQMQ,
  title={MoVQ: Modulating Quantized Vectors for High-Fidelity Image Generation},
  author={Chuanxia Zheng and Long Tung Vuong and Jianfei Cai and Dinh Q. Phung},
  journal={ArXiv},
  year={2022},
  volume={abs/2209.09002},
  url={https://api.semanticscholar.org/CorpusID:252367709}
}

@article{Tian2024VisualAM,
  title={Visual Autoregressive Modeling: Scalable Image Generation via Next-Scale Prediction},
  author={Keyu Tian and Yi Jiang and Zehuan Yuan and Bingyue Peng and Liwei Wang},
  journal={ArXiv},
  year={2024},
  volume={abs/2404.02905},
  url={https://api.semanticscholar.org/CorpusID:268876071}
}

@misc{zhang2025quantizethenrectifyefficientvqvaetraining,
      title={Quantize-then-Rectify: Efficient VQ-VAE Training}, 
      author={Borui Zhang and Qihang Rao and Wenzhao Zheng and Jie Zhou and Jiwen Lu},
      year={2025},
      eprint={2507.10547},
      archivePrefix={arXiv},
      primaryClass={cs.CV},
      url={https://arxiv.org/abs/2507.10547}, 
}

@article{Esser2020TamingTF,
  title={Taming Transformers for High-Resolution Image Synthesis},
  author={Patrick Esser and Robin Rombach and Bj{\"o}rn Ommer},
  journal={2021 IEEE/CVF Conference on Computer Vision and Pattern Recognition (CVPR)},
  year={2020},
  pages={12868-12878},
  url={https://api.semanticscholar.org/CorpusID:229297973}
}

@article{Han2024InfinitySB,
  title={Infinity: Scaling Bitwise AutoRegressive Modeling for High-Resolution Image Synthesis},
  author={Jian Han and Jinlai Liu and Yi Jiang and Bin Yan and Yuqi Zhang and Zehuan Yuan and Bingyue Peng and Xiaobing Liu},
  journal={ArXiv},
  year={2024},
  volume={abs/2412.04431},
  url={https://api.semanticscholar.org/CorpusID:274515181}
}

@inproceedings{esser2021taming,
  title={Taming transformers for high-resolution image synthesis},
  author={Esser, Patrick and Rombach, Robin and Ommer, Bjorn},
  booktitle={Proceedings of the IEEE/CVF conference on computer vision and pattern recognition},
  pages={12873--12883},
  year={2021}
}

@article{Sargent2025FlowTT,
  title={Flow to the Mode: Mode-Seeking Diffusion Autoencoders for State-of-the-Art Image Tokenization},
  author={Kyle Sargent and Kyle Hsu and Justin Johnson and Fei-Fei Li and Jiajun Wu},
  journal={ArXiv},
  year={2025},
  volume={abs/2503.11056},
  url={https://api.semanticscholar.org/CorpusID:277043290}
}

@article{xiong2025gigatok,
  title={Gigatok: Scaling visual tokenizers to 3 billion parameters for autoregressive image generation},
  author={Xiong, Tianwei and Liew, Jun Hao and Huang, Zilong and Feng, Jiashi and Liu, Xihui},
  journal={arXiv preprint arXiv:2504.08736},
  year={2025}
}

@article{wang2024image,
  title={Image understanding makes for a good tokenizer for image generation},
  author={Wang, Luting and Zhao, Yang and Zhang, Zijian and Feng, Jiashi and Liu, Si and Kang, Bingyi},
  journal={Advances in Neural Information Processing Systems},
  volume={37},
  pages={31015--31035},
  year={2024}
}

@article{hansen2025learnings,
  title={Learnings from scaling visual tokenizers for reconstruction and generation},
  author={Hansen-Estruch, Philippe and Yan, David and Chung, Ching-Yao and Zohar, Orr and Wang, Jialiang and Hou, Tingbo and Xu, Tao and Vishwanath, Sriram and Vajda, Peter and Chen, Xinlei},
  journal={arXiv preprint arXiv:2501.09755},
  year={2025}
}

@inproceedings{zhang2023regularized,
  title={Regularized vector quantization for tokenized image synthesis},
  author={Zhang, Jiahui and Zhan, Fangneng and Theobalt, Christian and Lu, Shijian},
  booktitle={Proceedings of the IEEE/CVF Conference on Computer Vision and Pattern Recognition},
  pages={18467--18476},
  year={2023}
}

@inproceedings{chiu2022self,
  title={Self-supervised learning with random-projection quantizer for speech recognition},
  author={Chiu, Chung-Cheng and Qin, James and Zhang, Yu and Yu, Jiahui and Wu, Yonghui},
  booktitle={International Conference on Machine Learning},
  pages={3915--3924},
  year={2022},
  organization={PMLR}
}

@article{yu2021vector,
  title={Vector-quantized image modeling with improved vqgan},
  author={Yu, Jiahui and Li, Xin and Koh, Jing Yu and Zhang, Han and Pang, Ruoming and Qin, James and Ku, Alexander and Xu, Yuanzhong and Baldridge, Jason and Wu, Yonghui},
  journal={arXiv preprint arXiv:2110.04627},
  year={2021}
}

@article{goswami2024hypervq,
  title={Hypervq: Mlr-based vector quantization in hyperbolic space},
  author={Goswami, Nabarun and Mukuta, Yusuke and Harada, Tatsuya},
  journal={arXiv preprint arXiv:2403.13015},
  year={2024}
}

@article{gautam2023soft,
  title={Soft convex quantization: Revisiting vector quantization with convex optimization},
  author={Gautam, Tanmay and Pryzant, Reid and Yang, Ziyi and Zhu, Chenguang and Sojoudi, Somayeh},
  journal={arXiv preprint arXiv:2310.03004},
  year={2023}
}

@article{takida2022sq,
  title={Sq-vae: Variational bayes on discrete representation with self-annealed stochastic quantization},
  author={Takida, Yuhta and Shibuya, Takashi and Liao, WeiHsiang and Lai, Chieh-Hsin and Ohmura, Junki and Uesaka, Toshimitsu and Murata, Naoki and Takahashi, Shusuke and Kumakura, Toshiyuki and Mitsufuji, Yuki},
  journal={arXiv preprint arXiv:2205.07547},
  year={2022}
}

@inproceedings{huh2023straightening,
  title={Straightening out the straight-through estimator: Overcoming optimization challenges in vector quantized networks},
  author={Huh, Minyoung and Cheung, Brian and Agrawal, Pulkit and Isola, Phillip},
  booktitle={International Conference on Machine Learning},
  pages={14096--14113},
  year={2023},
  organization={PMLR}
}

@article{mentzer2023finite,
  title={Finite scalar quantization: Vq-vae made simple},
  author={Mentzer, Fabian and Minnen, David and Agustsson, Eirikur and Tschannen, Michael},
  journal={arXiv preprint arXiv:2309.15505},
  year={2023}
}

@article{Song2020ScoreBasedGM,
  title={Score-Based Generative Modeling through Stochastic Differential Equations},
  author={Yang Song and Jascha Narain Sohl-Dickstein and Diederik P. Kingma and Abhishek Kumar and Stefano Ermon and Ben Poole},
  journal={ArXiv},
  year={2020},
  volume={abs/2011.13456},
  url={https://api.semanticscholar.org/CorpusID:227209335}
}

@article{Peebles2022ScalableDM,
  title={Scalable Diffusion Models with Transformers},
  author={William S. Peebles and Saining Xie},
  journal={2023 IEEE/CVF International Conference on Computer Vision (ICCV)},
  year={2022},
  pages={4172-4182},
  url={https://api.semanticscholar.org/CorpusID:254854389}
}

@article{Maaten2008VisualizingDU,
  title={Visualizing Data using t-SNE},
  author={Laurens van der Maaten and Geoffrey E. Hinton},
  journal={Journal of Machine Learning Research},
  year={2008},
  volume={9},
  pages={2579-2605},
  url={https://api.semanticscholar.org/CorpusID:5855042}
}

@article{wang2025bridging,
  title={Bridging continuous and discrete tokens for autoregressive visual generation},
  author={Wang, Yuqing and Lin, Zhijie and Teng, Yao and Zhu, Yuanzhi and Ren, Shuhuai and Feng, Jiashi and Liu, Xihui},
  journal={arXiv preprint arXiv:2503.16430},
  year={2025}
}

@article{Esser2024ScalingRF,
  title={Scaling Rectified Flow Transformers for High-Resolution Image Synthesis},
  author={Patrick Esser and Sumith Kulal and A. Blattmann and Rahim Entezari and Jonas Muller and Harry Saini and Yam Levi and Dominik Lorenz and Axel Sauer and Frederic Boesel and Dustin Podell and Tim Dockhorn and Zion English and Kyle Lacey and Alex Goodwin and Yannik Marek and Robin Rombach},
  journal={ArXiv},
  year={2024},
  volume={abs/2403.03206},
  url={https://api.semanticscholar.org/CorpusID:268247980}
}

@inproceedings{chang2022maskgit,
  title={Maskgit: Masked generative image transformer},
  author={Chang, Huiwen and Zhang, Han and Jiang, Lu and Liu, Ce and Freeman, William T},
  booktitle={Proceedings of the IEEE/CVF conference on computer vision and pattern recognition},
  pages={11315--11325},
  year={2022}
}

@article{sun2024autoregressive,
  title={Autoregressive model beats diffusion: Llama for scalable image generation},
  author={Sun, Peize and Jiang, Yi and Chen, Shoufa and Zhang, Shilong and Peng, Bingyue and Luo, Ping and Yuan, Zehuan},
  journal={arXiv preprint arXiv:2406.06525},
  year={2024}
}

@inproceedings{Shi2024ScalableIT,
  title={Scalable Image Tokenization with Index Backpropagation Quantization},
  author={Fengyuan Shi and Zhuoyan Luo and Yixiao Ge and Yujiu Yang and Ying Shan and Limin Wang},
  year={2024},
  url={https://api.semanticscholar.org/CorpusID:274445794}
}

@article{Fifty2024RestructuringVQ,
  title={Restructuring Vector Quantization with the Rotation Trick},
  author={Christopher Fifty and Ronald G. Junkins and Dennis Duan and Aniketh Iger and Jerry Liu and Ehsan Amid and Sebastian Thrun and Christopher R'e},
  journal={ArXiv},
  year={2024},
  volume={abs/2410.06424},
  url={https://api.semanticscholar.org/CorpusID:273229218}
}

@article{Huang2023TowardsAI,
  title={Towards Accurate Image Coding: Improved Autoregressive Image Generation with Dynamic Vector Quantization},
  author={Mengqi Huang and Zhendong Mao and Zhuowei Chen and Yongdong Zhang},
  journal={2023 IEEE/CVF Conference on Computer Vision and Pattern Recognition (CVPR)},
  year={2023},
  pages={22596-22605},
  url={https://api.semanticscholar.org/CorpusID:258823089}
}

@article{Zhu2024ScalingTC,
  title={Scaling the Codebook Size of VQGAN to 100,000 with a Utilization Rate of 99\%},
  author={Lei Zhu and Fangyun Wei and Yanye Lu and Dong Chen},
  journal={ArXiv},
  year={2024},
  volume={abs/2406.11837},
  url={https://api.semanticscholar.org/CorpusID:270560634}
}

@article{Mentzer2020HighFidelityGI,
  title={High-Fidelity Generative Image Compression},
  author={Fabian Mentzer and George Toderici and Michael Tschannen and Eirikur Agustsson},
  journal={ArXiv},
  year={2020},
  volume={abs/2006.09965},
  url={https://api.semanticscholar.org/CorpusID:219721015}
}

@article{Havasi2018MinimalRC,
  title={Minimal Random Code Learning: Getting Bits Back from Compressed Model Parameters},
  author={Marton Havasi and Robert Peharz and Jos{\'e} Miguel Hern{\'a}ndez-Lobato},
  journal={ArXiv},
  year={2018},
  volume={abs/1810.00440},
  url={https://api.semanticscholar.org/CorpusID:52901536}
}

@article{Kingma2014AdamAM,
  title={Adam: A Method for Stochastic Optimization},
  author={Diederik P. Kingma and Jimmy Ba},
  journal={CoRR},
  year={2014},
  volume={abs/1412.6980},
  url={https://api.semanticscholar.org/CorpusID:6628106}
}

@inproceedings{Vonderfecht2025LossyCW,
  title={Lossy compression with pretrained diffusion models},
  author={Liu, Feng and others},
  booktitle={International Conference on Learning Representations},
  volume={2025},
  pages={687--702},
  year={2025}
}

@article{Lin2023MinimalRC,
  title={Minimal Random Code Learning with Mean-KL Parameterization},
  author={Jihao Andreas Lin and Gergely Flamich and Jos{\'e} Miguel Hern{\'a}ndez-Lobato},
  journal={ArXiv},
  year={2023},
  volume={abs/2307.07816},
  url={https://api.semanticscholar.org/CorpusID:259936972}
}

@article{Flamich2020CompressingIB,
  title={Compressing Images by Encoding Their Latent Representations with Relative Entropy Coding},
  author={Gergely Flamich and Marton Havasi and Jos{\'e} Miguel Hern{\'a}ndez-Lobato},
  journal={ArXiv},
  year={2020},
  volume={abs/2010.01185},
  url={https://api.semanticscholar.org/CorpusID:222132826}
}

@article{Lee2022AutoregressiveIG,
  title={Autoregressive Image Generation using Residual Quantization},
  author={Doyup Lee and Chiheon Kim and Saehoon Kim and Minsu Cho and Wook-Shin Han},
  journal={2022 IEEE/CVF Conference on Computer Vision and Pattern Recognition (CVPR)},
  year={2022},
  pages={11513-11522},
  url={https://api.semanticscholar.org/CorpusID:247244535}
}

@inproceedings{Razavi2019GeneratingDH,
  title={Generating Diverse High-Fidelity Images with VQ-VAE-2},
  author={Ali Razavi and A{\"a}ron van den Oord and Oriol Vinyals},
  booktitle={Neural Information Processing Systems},
  year={2019},
  url={https://api.semanticscholar.org/CorpusID:173990382}
}

@article{Zhuang2025WeTokPD,
  title={WeTok: Powerful Discrete Tokenization for High-Fidelity Visual Reconstruction},
  author={Shaobin Zhuang and Yiwei Guo and Canmiao Fu and Zhipeng Huang and Zeyue Tian and Fangyikang Wang and Ying Zhang and Chen Li and Yali Wang},
  journal={ArXiv},
  year={2025},
  volume={abs/2508.05599},
  url={https://api.semanticscholar.org/CorpusID:280546113}
}

@article{Lin2026iFSQIF,
  title={iFSQ: Improving FSQ for Image Generation with 1 Line of Code},
  author={Bin Lin and Zongjian Li and Yuwei Niu and Kaixiong Gong and Yunyang Ge and Yunlong Lin and Mingzhe Zheng and Jianwei Zhang and Miles Yang and Zhao Zhong and Liefeng Bo and Li Yuan},
  journal={ArXiv},
  year={2026},
  volume={abs/2601.17124},
  url={https://api.semanticscholar.org/CorpusID:285050418}
}

@article{Wang2004ImageQA,
  title={Image quality assessment: from error visibility to structural similarity},
  author={Zhou Wang and Alan Conrad Bovik and Hamid R. Sheikh and Eero P. Simoncelli},
  journal={IEEE Transactions on Image Processing},
  year={2004},
  volume={13},
  pages={600-612},
  url={https://api.semanticscholar.org/CorpusID:207761262}
}

@article{Yu2021VectorquantizedIM,
  title={Vector-quantized Image Modeling with Improved VQGAN},
  author={Jiahui Yu and Xin Li and Jing Yu Koh and Han Zhang and Ruoming Pang and James Qin and Alexander Ku and Yuanzhong Xu and Jason Baldridge and Yonghui Wu},
  journal={ArXiv},
  year={2021},
  volume={abs/2110.04627},
  url={https://api.semanticscholar.org/CorpusID:238582653}
}

@article{Sun2024AutoregressiveMB,
  title={Autoregressive Model Beats Diffusion: Llama for Scalable Image Generation},
  author={Peize Sun and Yi Jiang and Shoufa Chen and Shilong Zhang and Bingyue Peng and Ping Luo and Zehuan Yuan},
  journal={ArXiv},
  year={2024},
  volume={abs/2406.06525},
  url={https://api.semanticscholar.org/CorpusID:270371603}
}

@inproceedings{Lin2014MicrosoftCC,
  title={Microsoft COCO: Common Objects in Context},
  author={Tsung-Yi Lin and Michael Maire and Serge J. Belongie and James Hays and Pietro Perona and Deva Ramanan and Piotr Doll{\'a}r and C. Lawrence Zitnick},
  booktitle={European Conference on Computer Vision},
  year={2014},
  url={https://api.semanticscholar.org/CorpusID:14113767}
}

@article{Shi2024TamingSV,
  title={Taming Scalable Visual Tokenizer for Autoregressive Image Generation},
  author={Fengyuan Shi and Zhuoyan Luo and Yixiao Ge and Yujiu Yang and Ying Shan and Limin Wang},
  journal={ArXiv},
  year={2024},
  volume={abs/2412.02692},
  url={https://api.semanticscholar.org/CorpusID:277772880}
}

@article{Deng2009ImageNetAL,
  title={ImageNet: A large-scale hierarchical image database},
  author={Jia Deng and Wei Dong and Richard Socher and Li-Jia Li and K. Li and Li Fei-Fei},
  journal={2009 IEEE Conference on Computer Vision and Pattern Recognition},
  year={2009},
  pages={248-255},
  url={https://api.semanticscholar.org/CorpusID:57246310}
}

@article{Salimans2016ImprovedTF,
  title={Improved Techniques for Training GANs},
  author={Tim Salimans and Ian J. Goodfellow and Wojciech Zaremba and Vicki Cheung and Alec Radford and Xi Chen},
  journal={ArXiv},
  year={2016},
  volume={abs/1606.03498},
  url={https://api.semanticscholar.org/CorpusID:1687220}
}

@article{Zhang2018TheUE,
  title={The Unreasonable Effectiveness of Deep Features as a Perceptual Metric},
  author={Richard Zhang and Phillip Isola and Alexei A. Efros and Eli Shechtman and Oliver Wang},
  journal={2018 IEEE/CVF Conference on Computer Vision and Pattern Recognition},
  year={2018},
  pages={586-595},
  url={https://api.semanticscholar.org/CorpusID:4766599}
}

@inproceedings{Heusel2017GANsTB,
  title={GANs Trained by a Two Time-Scale Update Rule Converge to a Local Nash Equilibrium},
  author={Martin Heusel and Hubert Ramsauer and Thomas Unterthiner and Bernhard Nessler and Sepp Hochreiter},
  booktitle={Neural Information Processing Systems},
  year={2017},
  url={https://api.semanticscholar.org/CorpusID:326772}
}

@article{Touvron2023Llama2O,
  title={Llama 2: Open Foundation and Fine-Tuned Chat Models},
  author={Hugo Touvron and Louis Martin and Kevin R. Stone and Peter Albert and Amjad Almahairi and Yasmine Babaei and Niko-lay Bashlykov and Soumya Batra and Prajjwal Bhargava and Shruti Bhosale and Daniel M. Bikel and Lukas Blecher and Cris-tian Cant{\'o}n Ferrer and Moya Chen and Guillem Cucurull and David Esiobu and Jude Fernandes and Jeremy Fu and Wenyin Fu and Brian Fuller and Cynthia Gao and Vedanuj Goswami and Naman Goyal and Anthony S. Hartshorn and Saghar Hosseini and Rui Hou and Hakan Inan and Marcin Kardas and Viktor Kerkez and Madian Khabsa and Isabel M. Kloumann and Artem Korenev and Punit Singh Koura and Marie-Anne Lachaux and Thibaut Lavril and Jenya Lee and Diana Liskovich and Yinghai Lu and Yuning Mao and Xavier Martinet and Todor Mihaylov and Pushkar Mishra and Igor Molybog and Yixin Nie and Andrew Poulton and Jeremy Reizenstein and Rashi Rungta and Kalyan Saladi and Alan Schelten and Ruan Silva and Eric Michael Smith and R. Subramanian and Xia Tan and Binh Tang and Ross Taylor and Adina Williams and Jian Xiang Kuan and Puxin Xu and Zhengxu Yan and Iliyan Zarov and Yuchen Zhang and Angela Fan and Melissa Hall Melanie Kambadur and Sharan Narang and Aur'elien Rodriguez and Robert Stojnic and Sergey Edunov and Thomas Scialom},
  journal={ArXiv},
  year={2023},
  volume={abs/2307.09288},
  url={https://api.semanticscholar.org/CorpusID:259950998}
}

@article{yu2023language,
  title={Language Model Beats Diffusion--Tokenizer is Key to Visual Generation},
  author={Yu, Lijun and Lezama, Jos{\'e} and Gundavarapu, Nitesh B and Versari, Luca and Sohn, Kihyuk and Minnen, David and Cheng, Yong and Birodkar, Vighnesh and Gupta, Agrim and Gu, Xiuye and others},
  journal={arXiv preprint arXiv:2310.05737},
  year={2023}
}

@inproceedings{rombach2022high,
  title={High-resolution image synthesis with latent diffusion models},
  author={Rombach, Robin and Blattmann, Andreas and Lorenz, Dominik and Esser, Patrick and Ommer, Bj{\"o}rn},
  booktitle={Proceedings of the IEEE/CVF conference on computer vision and pattern recognition},
  pages={10684--10695},
  year={2022}
}

@inproceedings{sonderby2017continuous,
  title={Continuous relaxation training of discrete latent variable image models},
  author={S{\o}nderby, Casper Kaae and Poole, Ben and Mnih, Andriy},
  booktitle={Beysian DeepLearning workshop, NIPS},
  volume={201},
  year={2017}
}

@article{roy2018theory,
  title={Theory and experiments on vector quantized autoencoders},
  author={Roy, Aurko and Vaswani, Ashish and Neelakantan, Arvind and Parmar, Niki},
  journal={arXiv preprint arXiv:1805.11063},
  year={2018}
}

@article{bengio2013estimating,
  title={Estimating or propagating gradients through stochastic neurons for conditional computation},
  author={Bengio, Yoshua and L{\'e}onard, Nicholas and Courville, Aaron},
  journal={arXiv preprint arXiv:1308.3432},
  year={2013}
}

@article{maddison2016concrete,
  title={The concrete distribution: A continuous relaxation of discrete random variables},
  author={Maddison, Chris J and Mnih, Andriy and Teh, Yee Whye},
  journal={arXiv preprint arXiv:1611.00712},
  year={2016}
}

@article{van2017neural,
  title={Neural discrete representation learning},
  author={Van Den Oord, Aaron and Vinyals, Oriol and others},
  journal={Advances in neural information processing systems},
  volume={30},
  year={2017}
}

@article{brezinski1996extrapolation,
  title={Extrapolation algorithms and Pad{\'e} approximations: a historical survey},
  author={Brezinski, Claude},
  journal={Applied numerical mathematics},
  volume={20},
  number={3},
  pages={299--318},
  year={1996},
  publisher={Elsevier}
}

@article{corless1996lambert,
  title={On the Lambert W function},
  author={Corless, Robert M and Gonnet, Gaston H and Hare, David EG and Jeffrey, David J and Knuth, Donald E},
  journal={Advances in Computational mathematics},
  volume={5},
  number={1},
  pages={329--359},
  year={1996},
  publisher={Springer}
}

@article{zhao2024image,
  title={Image and video tokenization with binary spherical quantization},
  author={Zhao, Yue and Xiong, Yuanjun and Kr{\"a}henb{\"u}hl, Philipp},
  journal={arXiv preprint arXiv:2406.07548},
  year={2024}
}

@article{townsend2019practical,
  title={Practical lossless compression with latent variables using bits back coding},
  author={Townsend, James and Bird, Tom and Barber, David},
  journal={arXiv preprint arXiv:1901.04866},
  year={2019}
}

@inproceedings{hinton1993keeping,
  title={Keeping the neural networks simple by minimizing the description length of the weights},
  author={Hinton, Geoffrey E and Van Camp, Drew},
  booktitle={Proceedings of the sixth annual conference on Computational learning theory},
  pages={5--13},
  year={1993}
}

@article{he2024accelerating,
  title={Accelerating relative entropy coding with space partitioning},
  author={He, Jiajun and Flamich, Gergely and Hern{\'a}ndez-Lobato, Jos{\'e} Miguel},
  journal={Advances in Neural Information Processing Systems},
  volume={37},
  pages={75791--75828},
  year={2024}
}

@article{Flamich2022FastRE,
  title={Fast Relative Entropy Coding with A* coding},
  author={Gergely Flamich and Stratis Markou and Jos'e Miguel Hern'andez-Lobato},
  journal={ArXiv},
  year={2022},
  volume={abs/2201.12857},
  url={https://api.semanticscholar.org/CorpusID:246430918}
}

@article{Theis2021AlgorithmsFT,
  title={Algorithms for the Communication of Samples},
  author={Lucas Theis and Noureldin Yosri},
  journal={ArXiv},
  year={2021},
  volume={abs/2110.12805},
  url={https://api.semanticscholar.org/CorpusID:239768821}
}

@article{havasi2018minimal,
  title={Minimal random code learning: Getting bits back from compressed model parameters},
  author={Havasi, Marton and Peharz, Robert and Hern{\'a}ndez-Lobato, Jos{\'e} Miguel},
  journal={arXiv preprint arXiv:1810.00440},
  year={2018}
}

@article{Luo2024OpenMAGVIT2AO,
  title={Open-MAGVIT2: An Open-Source Project Toward Democratizing Auto-regressive Visual Generation},
  author={Zhuoyan Luo and Fengyuan Shi and Yixiao Ge and Yujiu Yang and Limin Wang and Ying Shan},
  journal={ArXiv},
  year={2024},
  volume={abs/2409.04410},
  url={https://api.semanticscholar.org/CorpusID:272463752}
}

@article{li2018strong,
  title={Strong functional representation lemma and applications to coding theorems},
  author={Li, Cheuk Ting and El Gamal, Abbas},
  journal={IEEE Transactions on Information Theory},
  volume={64},
  number={11},
  pages={6967--6978},
  year={2018},
  publisher={IEEE}
}

@inproceedings{harsha2007communication,
  title={The communication complexity of correlation},
  author={Harsha, Prahladh and Jain, Rahul and McAllester, David and Radhakrishnan, Jaikumar},
  booktitle={Twenty-Second Annual IEEE Conference on Computational Complexity (CCC'07)},
  pages={10--23},
  year={2007},
  organization={IEEE}
}
\bibliographystyle{icml2026}

\newpage
\appendix
\onecolumn
\section{Proof of Main Results}
\label{app:proof}
\begin{lemma}
\label{thm:c}
    Given $R_i=D_{KL}(q(Z_i|X)||\mathcal{N}(0,1)),|\mu_i|,|\sigma_i|$ are upper-bounded. And therefore $c_1,c_2$ in Theorem~\ref{thm:ach}-\ref{thm:con} are also bounded:
    \begin{gather}
        |\mu_i| \le \sqrt{2R_i},\notag \\
        |\sigma_i| \le \sqrt{-\mathcal{W}_{-1}(-e^{-(2R_i+1)})}, \notag \\
        c_1 \le \sqrt{2R_i}\sqrt{-\mathcal{W}_{-1}(-e^{-(2R_i+1)})},\notag \\
        c_2 \le \sqrt{2R_i}+\sqrt{-\mathcal{W}_{-1}(-e^{-(2R_i+1)})}.\notag
    \end{gather}
\end{lemma}
\begin{proof}
First, we have
\begin{gather}
    \mu_i^2 + \sigma_i^2 -1 - \log \sigma_i^2 = 2 R_i.
\end{gather}
Notice that $\sigma_i^2-\log \sigma_i^2 \ge 0$. To maximize $\sigma_i$, we simply set $\mu_i=0$. Then we have $\sigma_i^2-\log \sigma_i^2 = 2R_i+1$. We can solve 
\begin{gather}
    \sigma_i^2 = -\mathcal{W}(e^{-(2R_i+1)}),
\end{gather}
where $\mathcal{W}(.)$ is Lambert $\mathcal{W}$ function. We select the larger solution 
\begin{gather}
    \sigma^2 = -\mathcal{W}_{-1}(e^{-(2R_i+1)}),
\end{gather}
where $\mathcal{W}_{-1}$ is the lower branch of Lambert $\mathcal{W}$ function.
To maximize $\mu_i$, we simply set $\sigma_i=1$. Then we have
\begin{gather}
    \mu_i^2 = 2 R_i,\mu_i = \sqrt{2R_i}.
\end{gather}
\end{proof}

\textbf{Theorem 3.1.} \textit{
    Denote the mean and standard deviation of $q(Z_i|X=x)$ as $\mu_i$ and $\sigma_i$, respectively. We assume $|\mu_i\sigma_i|\le c_1$ and $|\mu_i|+|\sigma_i|\le c_2$. Given a fixed $R_i = D_{KL}(q(Z_i|X)||\mathcal{N}(0,1))$, the probability of a quantization error $|\hat{z}_i - \mu_i| \geq \sigma_i$ decays doubly exponentially with the number of nats $t$ by which the codebook bitrate $\log K$ exceeds the bits-back coding rate. That is,}
    \begin{gather}        
        \textup{when } \log K = R_i + t,\notag \\\Pr\{|\hat{z}_i - \mu_i| \ge \sigma_i\} \le \exp{(-e^{t}\sqrt{\frac{2}{\pi}}e^{-c_1-0.5})}.
    \end{gather}
\begin{proof}
    Denote the cumulative distribution function (CDF) of $\mathcal{N}(0,1)$ as $\Phi$, and probability density function (PDF) of $\mathcal{N}(0,1)$ as $\phi$, then we need to consider the probability that no samples falls between $[\mu_i-\sigma_i, \mu_i+\sigma_i]$, which is
    \begin{gather}
        \Pr\{|\hat{z}_i - \mu_i| \ge \sigma_i\} = (1 - (\Phi(\mu_i + \sigma_i) - \Phi(\mu_i - \sigma_i)))^K. \label{eq:pr}
    \end{gather}
    Now we use the Bernoulli inequality, that $\forall y \in \mathbb{R}, 1 + y \le e^{y}$. Let $y=-(\Phi(\mu_i + \sigma_i) - \Phi(\mu_i - \sigma_i))$, we have
    \begin{gather}
        1 - (\Phi(\mu_i + \sigma_i) - \Phi(\mu_i - \sigma_i)) \le \exp{(-(\Phi(\mu_i + \sigma_i) - \Phi(\mu_i - \sigma_i)))}.\label{eq:br}
    \end{gather}
    Taking Eq.~\ref{eq:br} into Eq.~\ref{eq:pr}, we have
    \begin{align}        
        \Pr\{|\hat{z}_i - \mu_i| \ge \sigma_i\} &\le \exp{(-(\Phi(\mu_i + \sigma_i) - \Phi(\mu_i - \sigma_i)))}^K \\ \notag
        &= \exp{(-K \cdot (\Phi(\mu_i + \sigma_i) - \Phi(\mu_i - \sigma_i)))} \\ \notag
        &= \exp{(-K \cdot \int_{\mu_i - \sigma_i}^{\mu_i+\sigma_i}\phi(x)dx)}.
    \end{align}
    By integral mean value theorem, $\exists x' \in [\mu_i - \sigma_i, \mu_i + \sigma_i]$, such that 
    \begin{gather}
\int_{\mu_i - \sigma_i}^{\mu_i+\sigma_i}\phi(x)dx = 2\sigma_i \phi(x').
    \end{gather}
    And then we have 
    \begin{gather}
    \Pr\{|\hat{z}_i - \mu_i| \ge \sigma_i\} \le \exp{(-K\cdot 2\sigma_i\phi(x'))}.
    \end{gather}
    Next, we consider three cases: $\mu_i - \sigma_i \ge 0$, $\mu_i + \sigma_i \le 0$, and $\mu_i - \sigma_i \le 0 \le \mu_i + \sigma_i$.
    
    First, consider the case when $\mu_i - \sigma_i \ge 0$. Obviously we have $\phi(\mu_i + \sigma_i) \le \phi(x')$, and we have
    \begin{align}        
    \Pr\{|\hat{z}_i - \mu_i| \ge \sigma_i\} &\le \exp{(-K\cdot 2\sigma_i\phi(\mu_i+\sigma_i))} \notag \\ 
    &=\exp{(-K\cdot \sqrt{\frac{2}{\pi}}\sigma_ie^{-\frac{1}{2}(\mu_i+\sigma_i)^2})} \notag \\
    &=\exp{(-K\cdot \sqrt{\frac{2}{\pi}}e^{-\frac{1}{2}(\mu_i^2+\sigma_i^2-\log \sigma^2 -1.0 + 1.0)-\mu_i\sigma_i})} \notag \\
    &=\exp{(-K\cdot \sqrt{\frac{2}{\pi}}e^{-R_i - \mu_i\sigma_i-0.5})}.
    \end{align}
    Notice that as $\mu_i - \sigma_i \ge 0, \sigma_i > 0$, we must have $\mu_i\sigma_i > 0$, then 
    \begin{align}
    \Pr\{|\hat{z}_i - \mu_i| \ge \sigma_i\} &\le \exp{(-K\cdot \sqrt{\frac{2}{\pi}}e^{-R_i - |\mu_{\max}\sigma_{\max}|-0.5})}.
    \end{align}
    Similarly, we can show similar result for $\mu_i + \sigma_i \le 0$. Obviously we have $\phi(\mu_i - \sigma_i) \le \phi(x')$, and we have
    \begin{align}        
    \Pr\{|\hat{z}_i - \mu_i| \ge \sigma_i\} &\le \exp{(-K\cdot 2\sigma_i\phi(\mu_i-\sigma_i))} \notag \\ 
    &=\exp{(-K\cdot \sqrt{\frac{2}{\pi}}\sigma_ie^{-\frac{1}{2}(\mu_i+\sigma_i)^2})} \notag \\
    &=\exp{(-K\cdot \sqrt{\frac{2}{\pi}}e^{-\frac{1}{2}(\mu_i^2+\sigma_i^2-\log \sigma^2 -1.0 + 1.0)+\mu_i\sigma_i})} \notag \\
    &=\exp{(-K\cdot \sqrt{\frac{2}{\pi}}e^{-R_i + \mu_i\sigma_i-0.5})} \notag \\
    &\le \exp{(-K\cdot \sqrt{\frac{2}{\pi}}e^{-R_i - |\mu_{\max}\sigma_{\max}|-0.5})}
    \end{align}
    Now, consider the case when $\mu_i - \sigma_i < 0 < \mu_i + \sigma_i$, obviously we must have either $\phi(\mu_i - \sigma_i) \le \phi(x')$, or $\phi(\mu_i + \sigma_i) \le \phi(x')$. If $\phi(\mu_i + \sigma_i) \le \phi(x')$, then the result is the same as $\mu - \sigma_i \ge 0$. If $\phi(\mu_i - \sigma_i) \le \phi(x')$, then the result is the same as $\mu + \sigma_i \le 0$. 
    
    Therefore, for all $\mu_i,\sigma_i$, we have 
    \begin{align}
         \Pr\{|\hat{z}_i - \mu_i| \ge \sigma_i\} &\le \exp{(-K\cdot \sqrt{\frac{2}{\pi}}e^{-R_i - |\mu_{\max}\sigma_{\max}|-0.5})}.
    \end{align}
    
    Taking the value of $K$ in, we have the result
    \begin{align}
        \Pr\{|\hat{z}_i - \mu_i| \ge \sigma_i\} &\le \exp{(-K\cdot \sqrt{\frac{2}{\pi}}e^{-R_i-|\mu_{\max}\sigma_{\max}|-0.5})} \notag \\
        &=\exp{(-e^{t}\cdot \sqrt{\frac{2}{\pi}}e^{-|\mu_{\max}\sigma_{\max}|-0.5})}. \notag \\
        &=\exp{(-e^{t}\cdot \sqrt{\frac{2}{\pi}}e^{-c_1-0.5})}.
    \end{align}
\end{proof}

\textbf{Theorem 3.2.} \textit{
    Following Theorem~\ref{thm:ach}, the probability of a quantization error $|\hat{z}_i - \mu_i| \ge \sigma_i$ increases exponentially with the number of nats $t$ by which the codebook bitrate $\log K$ falls below the bits-back coding rate. That is,}
    \begin{gather}        
        \textup{when } \log K = R_i - t, \notag \\ 
        \Pr\{|\hat{z}_i - \mu_i| \ge \sigma_i\} \ge 1 - e^{-t}\sqrt{\frac{2}{\pi}}e^{0.5c_2^2-0.5}.
    \end{gather}
\begin{proof}
    Similar to the proof of Theorem.~\ref{thm:ach}, we have
    \begin{gather}
        \Pr(|\hat{z}_i - \mu_i|\ge \sigma_i) = (1 - (\Phi(\mu_i + \sigma_i) - \Phi(\mu_i - \sigma_i)))^K.
    \end{gather}
    Now we use an inequality, that $\forall y \in (0,1), K \in \mathbb{N}, K \ge 1, (1-y)^K \ge 1 - Ky$. This is due to the fact that $(1-y)^K$ is convex in $(0,1)$, and $1 - Ky$ is tangent line at $y=0$. With this inequality, we have 
    \begin{gather}
        (1 - (\Phi(\mu_i + \sigma_i) - \Phi(\mu_i - \sigma_i)))^K \ge 1 - K(\Phi(\mu_i + \sigma_i) - \Phi(\mu_i - \sigma_i)).
    \end{gather}
    Again, we can use integral mean value theorem, and find out that when $\mu_i - \sigma_i \ge 0$,
    \begin{align}
        \Pr(|\hat{z}_i - \mu_i| \ge \sigma_i) &\ge 1 - K(\Phi(\mu_i + \sigma_i) - \Phi(\mu_i - \sigma_i)) \notag \\
        & \ge 1 - K 2\sigma_i \phi(\mu_i - \sigma_i) \notag \\
        &=1 - K \sqrt{\frac{2}{\pi}}\sigma_ie^{-\frac{1}{2}(x_i-\sigma_i)^2} \notag \\
        &=1 - K\sqrt{\frac{2}{\pi}}e^{-\frac{1}{2}(x_i^2 + \sigma_i^2 - \log \sigma_i^2 -1.0) + |\mu_i\sigma_i|-0.5} \notag \\
        &=1 - K\sqrt{\frac{2}{\pi}}e^{-R_i + |\mu_i\sigma_i|-0.5} \notag \\
        &\ge 1 - K e^{-R_i} \sqrt{\frac{2}{\pi}}e^{0.5(\mu_i+\sigma_i)^2-0.5}
    \end{align}
    Similar results can be obtained for $\mu_i + \sigma_i \le 0$. For the case that $\mu_i - \sigma_i \le 0 \le \mu_i + \sigma_i$, we have 
    \begin{align}
        \Pr(|\hat{z}_i - \mu_i| \ge \sigma_i) &\ge 1 - K(\Phi(\mu_i + \sigma_i) - \Phi(\mu_i - \sigma_i)) \notag \\
        & \ge 1 - K 2\sigma_i \phi(0) \notag \\
        &=1 - K \sqrt{\frac{2}{\pi}}\sigma_ie^{-\frac{1}{2}(0)^2} \notag \\
        &= 1 - K\sqrt{\frac{2}{\pi}}e^{-\frac{1}{2}(\mu_i^2 + \sigma_i^2 - \log \sigma_i^2 -1.0) -0.5 + 0.5(\mu_i^2 + \sigma_i^2)} \notag \\
        &=1 - K\sqrt{\frac{2}{\pi}}e^{-R_i + 0.5(\mu_i^2 + \sigma_i^2) + |\mu_i\sigma_i|-0.5} \notag \\
        & = 1 - K e^{-R_i} \sqrt{\frac{2}{\pi}}e^{0.5(\mu_i+\sigma_i)^2-0.5}
    \end{align}
    Taking the value of $K = e^{R_i - t}$ , we have the result
    \begin{align}
        \Pr(|\hat{z}_i - \mu_i| \ge \sigma_i) &\ge 1 - e^{R_i - t-R_i} \sqrt{\frac{2}{\pi}}e^{0.5(\mu_i+\sigma_i)^2-0.5} \notag \\
        &\ge 1 - e^{-t}\sqrt{\frac{2}{\pi}}e^{0.5(|\mu_i|+|\sigma_i|)^2-0.5} \notag \\
        &\ge 1 - e^{-t}\sqrt{\frac{2}{\pi}}e^{0.5c_2^2-0.5}.
    \end{align}
\end{proof}
\section{Stable Mean-KL Parametrization}
\label{app:smkl}
We investigate an alternative to TDC, namely the Mean-KL parametrization \citep{Lin2023MinimalRC}, which is considered to be easier to train than TDC since it does not require the construction of an empirical $R(\lambda)$ model.

\subsection{Mean-KL Parametrization}
The Mean-KL parametrization \citep{Lin2023MinimalRC} supports codebook dimension $m > 1$. Its neural network output consists of two $m$-dimensional tensors, $\hat{\gamma}_{i:i+m}$ and $\tau_{i:i+m}$, which allocate the $R_{i:i+m}$ target $\log K$ across the $m$ dimensions and determine the mean, respectively. More specifically, the Mean-KL parametrization determines the mean $\mu_{i:i+m}$ and variance $\sigma^2_{i:i+m}$ as follows, where $\mathcal{W}(\cdot)$ denotes the principal branch of the Lambert $\mathcal{W}$ function:
\begin{gather}
    \gamma_{i:i+m} = \textrm{Softmax}(\hat{\gamma}_{i:i+m}),\notag \\
    \kappa_{i:i+m} = \gamma_{i:i+m}K, \notag \\
    \mu_{1:m} = \sqrt{2\kappa_{i:i+m}}\textrm{tanh}(\tau_{i:i+m}),\notag \\
    \sigma_{i:i+m}^2 = -\mathcal{W}(-\exp(\mu_{i:i+m}^2 - 2\kappa_{i:i+m}-1.0)).
\end{gather}
The Mean-KL parametrization is designed for model compression. When directly applied to Gaussian VAEs, two typical cases may arise, as shown in Table~\ref{tab:nan}, both of which can result in a not-a-number (NaN) error in floating-point computations.
\begin{table}[htb]
\caption{Two typical types of NaN in Mean-KL parametrization.}
\label{tab:nan}
\centering
\begin{tabular}{@{}ccc@{}}
\toprule
$\mu_i$ & $\kappa_i$ & $\sigma^2_i$ \\ \midrule
-2.7286 & 3.7227 & NaN \\
0.0013 & 9.1458$\times10^{-7}$ & NaN \\ \bottomrule
\end{tabular}
\end{table}
\subsection{Stable Mean-KL Parametrization}
It is evident that the two types of NaN errors are caused by excessively large values of $|\mu_i|$ and excessively small values of $\kappa_i$, respectively. To address this numerical issue, we propose the Stable Mean-KL parametrization, which introduces two regularization parameters, $r_1 = 0.1$ and $r_2 = 0.01$. The parameter $r_1$ ensures that each $\kappa_i \geq r_1 / m$, while $r_2$ shrinks $\mu_i$ towards $0$.
\begin{gather}
    \kappa_{i:i+m} = \gamma_{i:i+m}(K - r_1) + r_1/m, \notag \\
    \mu_{1:m} = \sqrt{2\kappa_{i:i+m}}\textrm{tanh}(\tau_{i:i+m})(1 - r_2),
\end{gather}
\subsection{Results of Stable Mean-KL Parametrization}
In Table~\ref{tab:rsmkl}, we present the results of the Stable Mean-KL parametrization. For UNet-based models, Stable Mean-KL achieves performance comparable to TDC. However, for ViT-based models, Stable Mean-KL performs significantly worse than TDC. Since Stable Mean-KL does not consistently outperform TDC, we choose to use TDC for the final model. Nonetheless, if only UNet-based models are required, Stable Mean-KL can be an effective alternative to TDC, as it does not require an empirical $R(\lambda)$ model and is significantly simpler to train.

\begin{table}[thb]
\caption{Quantitative results on ImageNet validation dataset comparing GQ with different constraint.}
\label{tab:rsmkl}
\centering
\begin{tabular}{@{}lccccccccc@{}}
\toprule
\multirow{2}{*}{Method} & \multicolumn{1}{c}{\multirow{2}{*}{bpp ($K\times N$)}} & \multicolumn{4}{c}{UNet based} & \multicolumn{4}{c}{ViT based} \\ \cmidrule(lr){3-6}\cmidrule(lr){7-10}
 & & PSNR$\uparrow$ & LPIPS$\downarrow$ & SSIM$\uparrow$ & rFID$\downarrow$ & PSNR$\uparrow$ & LPIPS$\downarrow$ & SSIM$\uparrow$ & rFID$\downarrow$ \\ \midrule
GQ (Mean-KL) & \multirow{3}{*}{1.00 (2$^{\textup{16}}\times$4096)} & NaN & NaN & NaN & NaN & NaN & NaN & NaN & NaN \\
GQ (Stable Mean-KL) &  & 32.35 & 0.023 & 0.905 & 0.280 & 30.80 & 0.030 & 0.891 & 0.556 \\
GQ (TDC) & & 32.47 & 0.023 & 0.907 & 0.322 & 31.71 & 0.024 & 0.903 & 0.349 \\ \bottomrule
\end{tabular}
\end{table}

\section{Implementation Details}
\label{app:impl}
\subsection{Details of Training and Distortion Objective} We train all VQ-VAEs on the ImageNet validation dataset using $8\times$ H100 GPUs for approximately 24 hours. For UNet models, we train each model for 30 epochs using the Adam \citep{Kingma2014AdamAM} optimizer with a batch size of $128$ and a learning rate of $1 \times 10^{-4}$. For ViT models, we train each model for 40 epochs using the Adam optimizer with a batch size of $256$ and a learning rate of $4 \times 10^{-7}$.

All VQ-VAEs are trained using the following distortion objective, which corresponds to the classical VQ-GAN \citep{Esser2020TamingTF} objective employed in the Stable Diffusion VAE \citep{rombach2022high}.
\begin{gather}
    \Delta(X,g(z)) = \mathcal{L}_{MSE}(X,g(z)) + w_1\mathcal{L}_{LPIPS}(X,g(z)) + w_2\mathcal{L}_{GAN}(g(z)).
\end{gather}
Following the implementation of Stable Diffusion, we set $w_1 = 1.0$ and $w_2 = 0.75$ for UNet models. Consistent with the implementation of BSQ \citep{zhao2024image}, we set $w_1 = 0.1$ and $w_2 = 0.1$ for ViT models.

For the image generation model, we first train all VQ-VAEs using images of size $128 \times 128$, following the same settings as described above. Subsequently, we train the auto-regressive transformer for image generation using the implementation of IBQ \citep{Shi2024TamingSV} with a Llama-base transformer architecture. The transformer has a vocabulary size of $2^{16}$, $16$ layers, $16$ attention heads, and an embedding dimension of $1024$. We train the transformer for $100$ epochs using the Adam optimizer with a learning rate of $3 \times 10^{-4}$ and a batch size of $512$.

\subsection{Details of Hyper-parameters}
Below, we describe the implementation details along with the definition of hyperparameters for each method. In Table~\ref{tab:hyp}, we list the values of these hyperparameters for different bits-per-pixel (bpp) settings.

\textbf{VQGAN} \citep{van2017neural} We adopt the factorized codebook VQGAN variant following \citep{Zheng2022MoVQMQ}. For each codebook, we use a codebook size of $K = 2^{16}$ and a group dimension of $m = 16$. The number of codebooks $n$ varies depending on the bitrate. Additionally, we use a codebook loss weight of $\lambda = 1.0$ and a commitment loss weight of $\zeta = 0.25$.

\textbf{FSQ} \citep{mentzer2023finite} The only parameter of FSQ is the codebook list $l$, which represents the quantization level for each integer value. We set each unit value to $2^{4} = 16$, and populate $l$ with the appropriate number of $16$s according to the desired bitrate.

\textbf{LFQ} \citep{yu2023language} For LFQ at $0.25$ bpp, we follow the original paper and split a large codebook of size $2^{16}$ into $n=2$ smaller codebooks, each with $K = 2^8$ and a codebook dimension of $m = 8$. We use an entropy loss weight of $\lambda = 0.1$ and a commitment loss weight of $\zeta = 0.025$.

\textbf{BSQ} \citep{zhao2024image} We fix the size of each BSQ codebook to $2^1$, with a group dimension of $m = 1$, and vary the number of codebooks $n$ according to the desired bitrate. For the entropy penalization parameter, we set $\lambda = 0.1$, following the official implementation.

\textbf{GQ} We use a fixed codebook size of $K = 2^{16}$. We adopt fixed $\alpha=0.5,\beta=1.01$. We adopt $m=16,8,4,\omega=2.0,2.0,0.0$ for bpp$=0.25,0.5,1.0$ respectively.

\begin{table}[thb]
\caption{Details of Hyper-parameter values.}
\label{tab:hyp}
\centering
\begin{tabular}{@{}lcc@{}}
\toprule
 & bpp & Hyper-parameters \\ \midrule
\multirow{3}{*}{VQ} & 0.25 & $K=2^{16},n=1,m=16,\lambda=1.0,\zeta=0.25$ \\
 & 0.50 & $K=2^{16},n=2,m=16,\lambda=1.0,\zeta=0.25$ \\
 & 1.00 & $K=2^{16},n=4,m=16,\lambda=1.0,\zeta=0.25$ \\ \midrule
\multirow{3}{*}{FSQ} & 0.25 & $l=\{16,16,16,16\}$ \\
 & 0.50 & $l=\{16,16,16,16,16,16,16,16\}$ \\
 & 1.00 & $l=\{16,16,16,16,16,16,16,16,16,16,16,16,16,16,16,16\}$ \\ \midrule
\multirow{3}{*}{LFQ} & 0.25 & $K=2^8,n=2,m=8,\lambda=0.1,\zeta=0.025$ \\
 & 0.50 & $K=2^8,n=4,m=8,\lambda=0.1,\zeta=0.025$ \\
 & 1.00 & $K=2^8,n=8,m=8,\lambda=0.1,\zeta=0.025$ \\ \midrule
\multirow{3}{*}{BSQ} & 0.25 & $K=2^1,n=16,m=1,\lambda=0.1$ \\
 & 0.50 & $K=2^1,n=32,m=1,\lambda=0.1$ \\
 & 1.00 & $K=2^1,n=64,m=1,\lambda=0.1$ \\ \midrule
\multirow{3}{*}{GQ} & 0.25 & $K=2^{16},n=1,m=16,\omega=2.0$ \\
 & 0.50 & $K=2^{16},n=2,m=8,\omega=2.0$ \\
 & 1.00 & $K=2^{16},n=4,m=4,\omega=0.0$ \\ \bottomrule
\end{tabular}
\end{table}

\section{Additional Quantitative Results}
\label{app:quant}

\subsection{Detail in Codebook Usage Hyper-parameter $\omega$}

Additionally, in Table~\ref{tab:ablbeta}, we show the effect of the codebook usage regularization parameter $\omega$. For high bitrates, such as $1.00$ bpp, regularization is not required; in other words, setting $\omega = 0.0$ yields the good enough codebook usage and rFID. For lower bitrates, such as $0.50$ bpp, $\omega = 0.0$ leads to codebook collapse, while $\omega = 2.0$ achieves the best codebook entropy and rFID.

\begin{table}[thb]
\caption{Ablation Study on regularization $\omega$.}
\label{tab:ablbeta}
\centering
\begin{tabular}{@{}lccccccc@{}}
\toprule
 bpp & $\omega$ & Codebook Usage$\uparrow$ & Codebook Entropy$\uparrow$ & PSNR$\uparrow$ & LPIPS$\downarrow$ & SSIM$\uparrow$ & rFID$\downarrow$ \\ \midrule
\multirow{4}{*}{0.50} & $0.0$ & 99.3\% & 14.96 & 30.00 & 0.044 & 0.873 & 0.783 \\
 & $1.0$ & 100.0\% & 15.14 & 30.35 & 0.040 & 0.877 & 0.589 \\ 
 & $2.0$ & 100.0\% & 15.22 & 30.17 & 0.039 & 0.875 & 0.492 \\ 
 & $4.0$ & 100.0\% & 14.81 & 28.08 & 0.061 & 0.846 & 1.269 \\ 
\midrule
\multirow{4}{*}{1.00} & $0.0$ & 100.0\% & 15.05 & 32.47 & 0.023 & 0.907 & 0.322 \\
 & $1.0$ & 100.0\% & 15.05 & 32.47 & 0.023 & 0.907 & 0.327 \\
 & $2.0$ & 100.0\% & 15.06 & 32.47 & 0.024 & 0.907 & 0.332 \\ 
 & $4.0$ & 100.0\% & 15.07 & 32.44 & 0.024 & 0.907 & 0.343 \\ \bottomrule
\end{tabular}
\end{table}

\begin{table}[thb]
\caption{Quantitative results on COCO 2017 dataset. \textbf{Bold}: best.}
\label{tab:rococo}
\centering
\begin{tabular}{@{}lccccccccc@{}}
\toprule
\multirow{2}{*}{Method} & \multicolumn{1}{c}{\multirow{2}{*}{bpp ($K\times N$)}} & \multicolumn{4}{c}{UNet based} & \multicolumn{4}{c}{ViT based} \\ \cmidrule(lr){3-6}\cmidrule(lr){7-10}
 & \multicolumn{1}{c}{} & \multicolumn{1}{c}{PSNR$\uparrow$} & \multicolumn{1}{c}{LPIPS$\downarrow$} & \multicolumn{1}{c}{SSIM$\uparrow$} & \multicolumn{1}{c}{rFID$\downarrow$} & \multicolumn{1}{c}{PSNR$\uparrow$} & \multicolumn{1}{c}{LPIPS$\downarrow$} & \multicolumn{1}{c}{SSIM$\uparrow$} & \multicolumn{1}{c}{rFID$\downarrow$} \\ \midrule
VQGAN & \multirow{5}{*}{0.25 (2$^{\textup{16}}\times$1024)} & 26.25 & 0.099 & 0.756 & 14.110 & 25.11 & 0.106 & 0.747 & 11.231 \\ 
FSQ &  & 26.01 & 0.072 & 0.767 & 5.451 & 25.85 & 0.112 & 0.765 & 11.213 \\
LFQ &  & 24.60 & 0.164 & 0.722 & 32.789 & 24.46 & 0.143 & 0.729 & 29.975 \\
BSQ &  & 25.29 & 0.085 & 0.763 & 5.803 & 26.15 & 0.082 & 0.798 & 7.034 \\
GQ (Ours) &  & \textbf{27.29} & \textbf{0.057} & \textbf{0.816} & \textbf{3.797} & \textbf{27.55} & \textbf{0.060} & \textbf{0.830} & \textbf{5.305} \\ \midrule
VQGAN & \multirow{5}{*}{0.50 (2$^{\textup{16}}\times$2048)} & 29.06 & 0.049 & 0.839 & 6.616 & 27.83 & 0.058 & 0.832 & 5.461 \\
FSQ &  & 29.08 & 0.043 & 0.855 & 4.008 & 28.51 & 0.053 & 0.851 & 5.390 \\
LFQ &  & 26.47 & 0.103 & 0.805 & 17.508 & 27.54 & 0.067 & 0.833 & 8.700 \\
BSQ &  & 27.58 & 0.057 & 0.844 & 4.465 & 28.19 & 0.049 & 0.858 & 4.587 \\
GQ (Ours) & & \textbf{30.14} & \textbf{0.037} & \textbf{0.877} & \textbf{3.116} & \textbf{30.18} & \textbf{0.034} & \textbf{0.887} & \textbf{3.616} \\ \midrule
VQGAN & \multirow{5}{*}{1.00 (2$^{\textup{16}}\times$4096)} & 31.97 & 0.024 & 0.901 & 3.455 & 31.07 & 0.029 & 0.904 & 3.494 \\
FSQ &  & 32.30 & 0.022 & \textbf{0.917} & 2.797 & 31.48 & 0.023 & 0.911 & 3.045 \\
LFQ &  & 28.16 & 0.072 & 0.845 & 11.121 & 26.36 & 0.103 & 0.794 & 20.381 \\
BSQ &  & 30.33 & 0.031 & 0.906 & 2.638 & 31.38 & 0.026 & \textbf{0.918} & 2.835 \\
GQ (Ours) & & \textbf{32.36} & \textbf{0.020} & 0.915 & \textbf{1.875} & \textbf{31.50} & \textbf{0.022} & 0.908 & \textbf{2.703} \\ \bottomrule
\end{tabular}
\end{table}

\begin{table}[htb]
\caption{The effect of quantization in pixel space.}
\label{tab:mean}
\centering
\begin{tabular}{@{}lccccc@{}}
\toprule
Latents & bits per latent & PSNR$\uparrow$ & LPIPS$\downarrow$ & SSIM$\uparrow$ & rFID$\downarrow$ \\ \midrule
$\mu_i = \mathbb{E}[Z_i|X]$ (posterior mean) & 16 bits & 32.92 & 0.020 & 0.913 & 0.46 \\
$z_i\sim q(Z_i|X)$ (Gaussian sample) & $D_{KL}(q(Z_i|X)||\mathcal{N}(0,1))=$ 4.26 bits  & 32.61 & 0.021 & 0.911 & 0.46 \\
$\hat{z}_i$ (GQ) & $\log_2 K=4$ bits & 32.11 & 0.023 & 0.906 & 0.414 \\
\bottomrule
\end{tabular}
\end{table}

\subsection{The Quantization Error in Pixel Space} Previously we examine the quantization error in latent space. We can further discuss the quantization error in pixel space given the decoder is smooth. More specifically, we have:

\textbf{Corollary 3. }\textit{Following the setting in Theorem~\ref{thm:ach}, and assuming the decoder $g(.)$ satisfy $|g(x_1)-g(x_2)| \le c_3 |x_1 - x_2|$, we have:}
\begin{gather}
    \textup{Pr}\{|g(\hat{z})-g(\mu)|\ge c_3\sigma_i\} \le \exp(-e^{t}\sqrt{\frac{2}{\pi}}e^{-c_1-0.5}).
\end{gather}
\textit{Proof. } As $|g(\hat{z}_i)-g(\mu_i)| \le c_3 |\hat{z}_i - \mu_i|$, we have $\textrm{Pr}\{|g(\hat{z}_i)-g(\mu_i)|\ge c_3\sigma_i\} \le \textrm{Pr}\{c_3 |\hat{z_i}-\mu_i|\ge c_3 \sigma_i\} = \textrm{Pr}\{|\hat{z_i}-\mu_i|\ge \sigma_i\}$.

We can see that theoretically, the quantization error can be magnified by the Lipschitz constant $c_3$. However, we note that this is not a significant issue in practice. As shown in the Table~\ref{tab:mean}, the actual loss of quality caused by GQ remains reasonable.

\subsection{Quantized Latent Visualization} 
\begin{figure}[thb]
    \centering    \includegraphics[width=\linewidth]{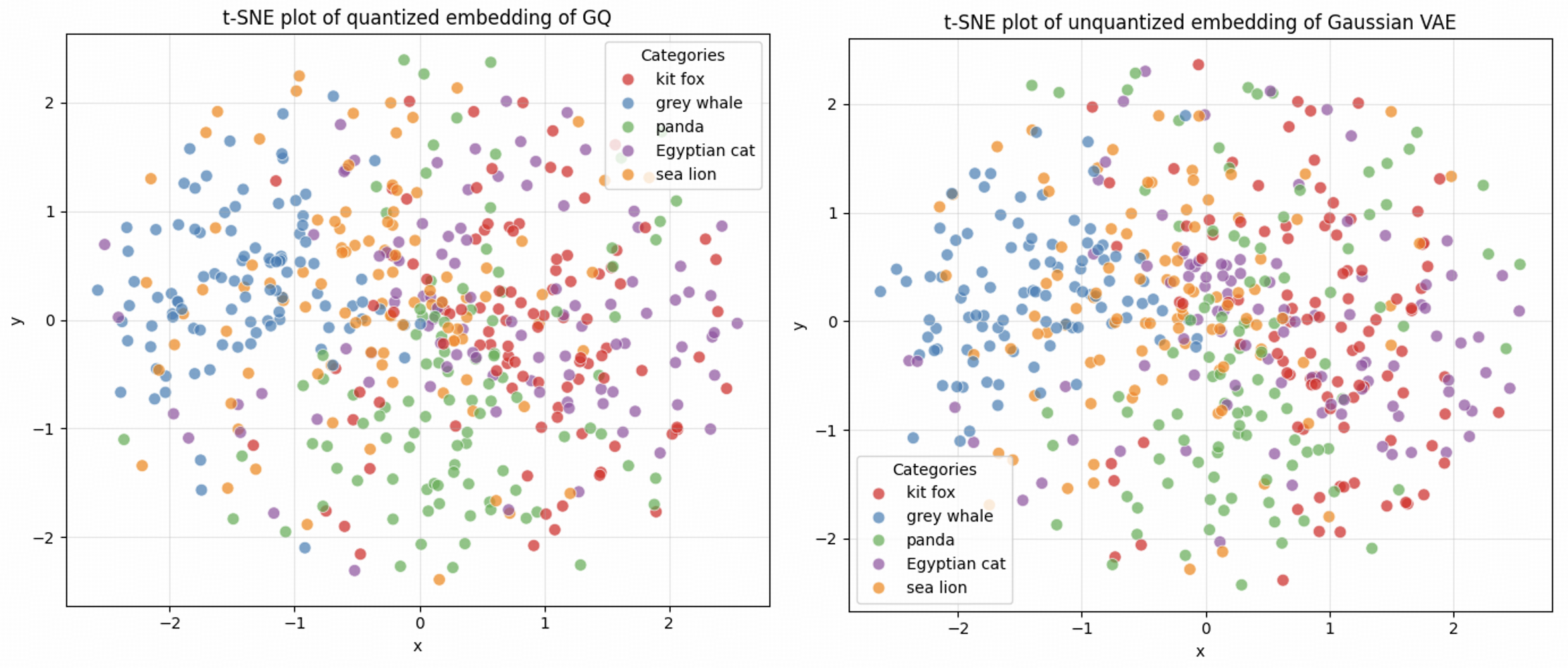}
    \caption{The t-NSE visualization of latent of GQ vs. unquantized Gaussian VAE. It is shown that the latent before and after quantization are quite similar.}
    \label{fig:tnse}
\end{figure}

In Figure~\ref{fig:tnse}, we show the t-NSE \citep{Maaten2008VisualizingDU} visualization of latent after GQ, using 5 subclass of ImageNet.

\subsection{More Generation Results} To better understand the generation performance, in Table~\ref{tab:gen-rate}, we present the auto-regressive generation result of GQ in different bitrate regime. And in Table~\ref{tab:gen-dataset}, we present the auto-regressive generation result of GQ in FFHQ dataset. It is shown that the advantage of GQ is consistent in different bitrate regime and datasets.

\begin{table}[]
\centering
\caption{The generation performance of GQ in different bitrate regime.}
\label{tab:gen-rate}
\begin{tabular}{@{}lccc@{}}
\toprule
Method & bpp (num of tokens) & gFID & IS \\ \midrule
TokenBridge & $0.1875 (2^{16}\times 256)$ & 8.29 & 188.05  \\
GQ (Ours) & $0.1875 (2^{16}\times 256)$ & 7.74	& 229.53 \\ \midrule
TokenBridge & $0.25 (2^{16}\times 256)$ & 7.82 & 198.24 \\
GQ (Ours) & $0.25 (2^{16}\times 256)$ &  7.67 & 230.79 \\ \bottomrule
\end{tabular}
\end{table}

\begin{table}[]
\caption{The generation performance of GQ with FFHQ dataset.}
\label{tab:gen-dataset}
\centering
\begin{tabular}{@{}lccc@{}}
\toprule
Method & bpp (num of tokens) & gFID & IS \\ \midrule
BSQ & $0.25 (2^{16}\times 256)$ & 5.48 & - \\
TokenBridge & $0.25 (2^{16}\times 256)$ & 7.15 & - \\
GQ (Ours) & $0.25 (2^{16}\times 256)$ & 5.09 & - \\ \bottomrule
\end{tabular}
\end{table}

\subsection{Prior-Posterior Mismatch} Sometimes the Gaussian VAE might suffer from prior-posterior mismatch. However, in our case, such mismatch is not severe. To illustrate this, we estimate the prior posterior mismatch by considering the relationship between $q(Z)$ and $\mathcal{N}(0,I)$. More specifically, we have
\begin{gather}
    D_{KL}(q(Z)||\mathcal{N}(0,1)) \approx \frac{1}{N}\sum_{i=1}^N (\log q(z^i) - \log \mathcal{N}(z^i|0,I)).
\end{gather}
Additionally, we can estimate the optimal bitrate without effected by prior posterior mismatch with a similar approximation:
\begin{gather}
    D_{KL}(q(Z|X)||p(Z)) \approx \frac{1}{N}\sum_{i=1}^N (\log q(z^i|X) - \log q(z^i)).
\end{gather}
We train a diffusion model to estimate $\log q(z^i)$ by using PF-ODE and Skilling-Hutchinson trace estimator (See Appendix D.2 of \citet{Song2020ScoreBasedGM}). We use the Gaussian VAE (w/ TDC) + DiT diffusion model and ImageNet validation dataset. The euler PF-ODE steps is set to 250 and the Skilling-Hutchinson number of sample is set to 1. The result is shown in Table~\ref{tab:mis}. The results show that the prior-posterior mismatch is only 0.00033 bits-per-pixel, accounting for approximately 0.1\% of the total bpp. Furthermore, the best bpp and the actual bpp show no significant difference on a scale of 0.01. This indicates that the "bitrate waste" caused by the prior-posterior mismatch is negligible, and the mismatch itself is not significant.

\begin{table}[]
\caption{The bitrate and prior-posterior mismatch.}
\label{tab:mis}
\centering
\begin{tabular}{@{}lc@{}}
\toprule
Divergence & bits-per-pixel \\ \midrule
bpp \textit{w.r.t.} $\mathcal{N}(0,I)$ ($D_{KL}(q(Z\|X)\|\|\mathcal{N}(0,1))$) & 0.25 \\
bpp \textit{w.r.t.} $q(Z)$ ($D_{KL}(q(Z\|X)\|\|q(Z))$) & 0.25 \\
prior posterior mismatch ($D_{KL}(q(Z)\|\|\mathcal{N}(0,1))$) & 0.000328 \\ \bottomrule
\end{tabular}
\end{table}
\subsection{Complexity}
\label{app:complex}
\begin{table}[thb]
\caption{The encoding and decoding overhead of GQ over Gaussian VAE.}
\label{tab:ct}
\centering
\begin{tabular}{@{}lcccc@{}}
\toprule
\multirow{2}{*}{Method} & \multicolumn{2}{c}{UNet based} \\ \cmidrule(lr){2-3}
 & Encoding FPS & Decoding FPS  \\ \midrule
Gaussian VAE & 104 & 64 \\
GQ (torch) & 12 & 61 \\
GQ (CUDA) & 79 & 61 \\ \bottomrule
\end{tabular}
\end{table}

As with FSQ and BSQ \citep{mentzer2023finite, zhao2024image}, our codebook can be generated on the fly by maintaining the same random number generator seed on both the encoder and decoder sides. Therefore, our GQ model has the same parameter size as the vanilla Gaussian VAE. In Table~\ref{tab:ct}, we compare the encoding and decoding frames per second (FPS) of the Gaussian VAE and GQ. We use $256 \times 256$ images with a batch size of $1$, and we report the wall clock time, meaning that the time required for loading data is included. The results show that the encoding FPS of GQ (implemented in PyTorch) is 12 on an H100 GPU, which is considerably slower than the 104 FPS achieved by the Gaussian VAE. On the other hand, GQ does not incur any decoding overhead.

To reduce the computational complexity of GQ, we implement GQ using a tailored CUDA kernel. Specifically, we follow the approach of \citet{Vonderfecht2025LossyCW}, with a key difference: we maintain the codebook, as our bottleneck is not codebook instantiation. Additionally, we avoid the creation of large buffer vectors by performing the summation over $m$ within the CUDA kernel instead of in PyTorch. With this approach, we achieve an encoding FPS of approximately 80, with negligible overhead compared to the Gaussian VAE. A detailed comparison between the PyTorch implementation and the CUDA implementation of GQ is provided below as {\fontfamily{qcr}\selectfont{GQ\_torch}} and {\fontfamily{qcr}\selectfont{GQ\_CUDA}}, respectively.

\begin{lstlisting}[language=Python]
def GQ_torch(mu, sigma, codebook, m, bs, K):
    # mu.shape = (bs, m)
    # sigma.shape = (bs, m)
    # codebook.shape = (K, m)

    # This step create (bs, m, K) vector, which is the performance bottleneck
    dist_m =((mu[:,None] - codebook[None])/sigma[:,None])**2
    dist = torch.sum(dist_m, dim=1) # sum over m dimension
    indices = torch.argmin(dist, dim=1) # argmax over K dimension
    zhat = torch.index_select(codebook, 0, indices) # select quantized results
    return indices, zhat
\end{lstlisting}

\begin{lstlisting}[language=Python]
def GQ_CUDA(mu, sigma, codebook, m, bs, K):
    dist = torch.zeros([bs, K])
    # need an extension wrapping and register the kernel into operator, we omit it in paper
    # see code appendix for details
    GQ_Kernel<<<bs * K / 256,256>>>(mu, sigma, codebook, dist, m, bs, K)
    
    indices = torch.argmin(dist, dim=1) # argmax over K dimension
    zhat = torch.index_select(codebook, 0, indices) # select quantized results
    return indices, zhat

__global__ void GQ_Kernel(
  const float* mu,
  const float* sigma,
  const float* codebook,
  float* dist,
  int64_t m,
  int64_t bs,
  int64_t K
) {
  int idx = blockIdx.x * blockDim.x + threadIdx.x;
  if (idx >= K * bs) return;
  int bi = idx / K;
  int ni = idx % K;
  float a = 0.0f;
  for (int i = 0; i < m; i++) {
      float b = (codebook[ni * m + i] - mu[bi * dim + i]) / sigma[bi * dim + i];
      a += b * b;
  }
  dist[idx] = a;
  return; 
}
\end{lstlisting}

\subsection{Asymptotic Complexity}
\label{app:asycomplex}

It is noteworthy that GQ with codebook dimension $m=1$, is asymptotically faster than reverse channel coding methods. This is because, for $m=1$, the GQ target in Eq.\ref{eq:qr} reduces to a quadratic form. In this case, it suffices to sort the scalar codebook $c_{1:K}$ in advance. Despite the sorting takes $\Omega(D_{KL}(q(Z_i|X)||\mathcal{N}(0,1)))$, the sorting is only need to be done once and can be amortized across dimension and dataset. Subsequently, the minimization in Eq.\ref{eq:qr} can be performed in $O(D_{\mathrm{KL}}(q(Z_i|X)||\mathcal{N}(0,1)))$ time using binary search. The details is shown in Algorithm~\ref{alg:gqbisect}.

On the other hand, most reverse channel coding methods require $O(2^{D_{\mathrm{KL}}(q(Z_i|X)||\mathcal{N}(0,1))})$ computational complexity \citep{havasi2018minimal, Flamich2020CompressingIB, Theis2021AlgorithmsFT}. A$^*$ coding \citep{Flamich2022FastRE} can achieve $O(D_{\infty}(q(Z_i|X)||\mathcal{N}(0,1)))$ encoding complexity, albeit at the cost of increased decoding complexity.

However, we note that this complexity advantage is not particularly meaningful in practice. This is because any auto-regressive generation model requires a softmax operation over the entire codebook, which has a complexity of $O(2^{D_{\mathrm{KL}}(q(Z_i|X)||\mathcal{N}(0,1))})$. In practice, only tractable codebook sizes, such as $2^{16}$ or $2^{18}$, are used.

\section{Additional Quantitative Results}

\subsection{Additional Qualitative Results and Failure Cases} 
\begin{figure}[thb]
    \centering
    \includegraphics[width=\linewidth]{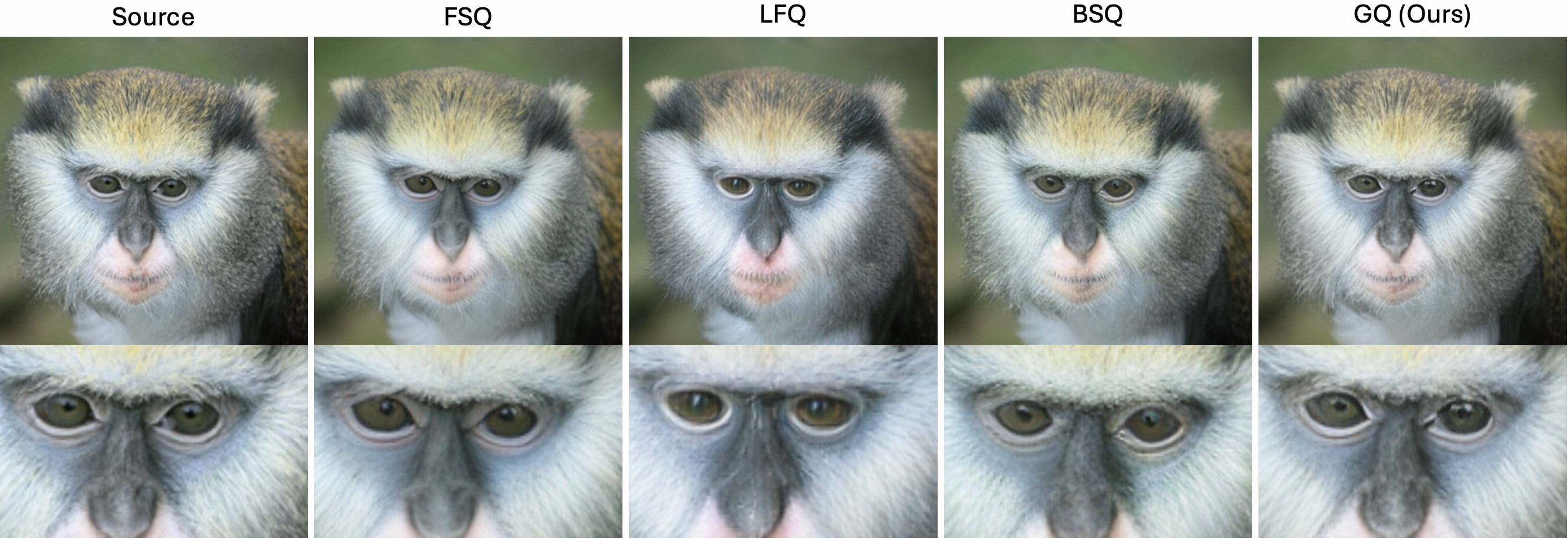}
    \includegraphics[width=\linewidth]{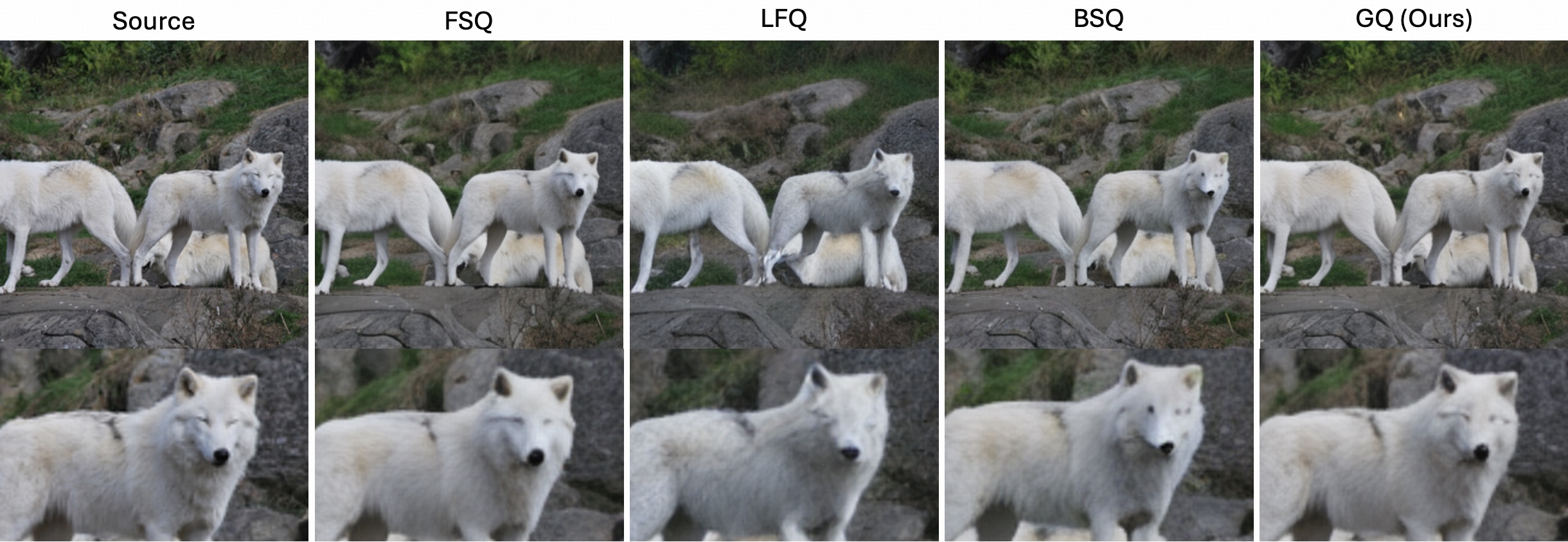}
    \includegraphics[width=\linewidth]{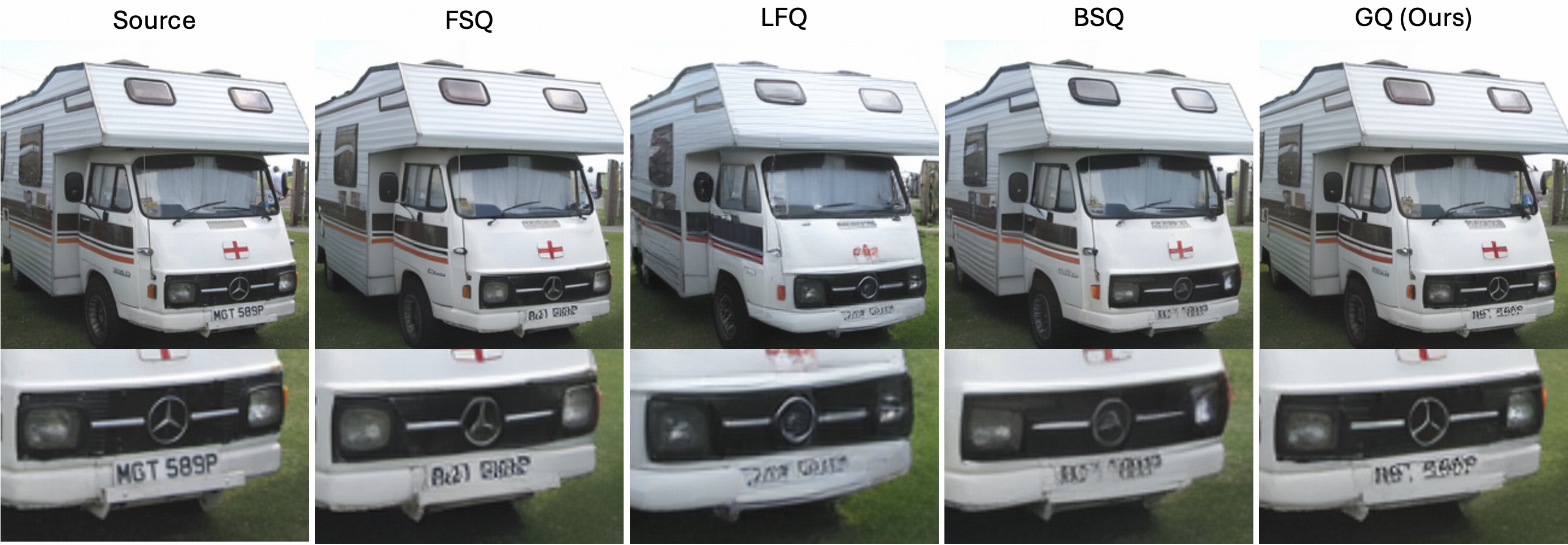}
    \caption{Qualitative results on ImageNet dataset and 0.25 bpp. None of these approaches correctly reconstructs the license plate.}
    \label{fig:qual2}
\end{figure}

In Figure~\ref{fig:qual2}, we present additional qualitative results showing that GQ achieves superior visual quality. However, we also note that none of the approaches is successful in reconstructing the license of the residential vehicle. The text content remains challenging for low bitrate VQ-VAEs.

\begin{minipage}[t]{0.48\textwidth}
\begin{algorithm}[H]
\caption{GQ (Argmax)}
\label{alg:gq_argmax}
\begin{algorithmic}[1]
\REQUIRE Codebook $c_{1:K}$ (sorted, $c_j \leq c_{j+1}$), $\mu_i$
\ENSURE Quantized value $\hat{z}_i$, index $j^*$
\STATE $T^* \gets \infty$
\FOR{$j = 1$ to $K$}
    \IF{$T^* \leq \|c_j - \mu_i\|$}
        \STATE $T^* \gets \|c_j - \mu_i\|$
        \STATE $j^* \gets j$
    \ENDIF
\ENDFOR
\STATE \textbf{return} $c_{j^*}, j^*$
\end{algorithmic}
\end{algorithm}
\end{minipage}
\hfill
\begin{minipage}[t]{0.48\textwidth}
\begin{algorithm}[H]
\caption{GQ (Bisect)}
\label{alg:gqbisect}
\begin{algorithmic}[1]
\REQUIRE Codebook $c_{1:K}$ (sorted, $c_j \leq c_{j+1}$), $\mu_i$
\ENSURE Quantized value $\hat{z}_i$, index $l$ or $r$
\STATE $l \gets 1$, $r \gets K$
\WHILE{$l + 1 < r$}
    \STATE $m \gets \lfloor (l + r) / 2 \rfloor$
    \IF{$c_m < \mu_i$}
        \STATE $l \gets m$
    \ELSE
        \STATE $r \gets m$
    \ENDIF
\ENDWHILE
\IF{$\|c_l - \mu_i\| < \|c_r - \mu_i\|$}
    \STATE \textbf{return} $c_l, l$
\ELSE
    \STATE \textbf{return} $c_r, r$
\ENDIF
\end{algorithmic}
\end{algorithm}
\end{minipage}

\end{document}